\documentclass[conference]{IEEEtran}
\usepackage{times}

\usepackage[numbers]{natbib}
\usepackage{multicol}
\usepackage{multirow}

\usepackage{graphicx,subcaption,ragged2e}
\usepackage{mathtools} 
\usepackage{amsmath} 
\usepackage{amssymb} 
\usepackage{colortbl} 
\usepackage[table]{xcolor} 
\definecolor{mycitecolor}{RGB}{71, 191, 38}
\definecolor{mylinkcolor}{RGB}{40, 115, 201}

\usepackage[bookmarks=true, colorlinks, citecolor=mycitecolor,linkcolor=mylinkcolor,urlcolor=mycitecolor ]{hyperref}

\usepackage{amsthm, setspace}
\usepackage{subcaption}
\usepackage{tikz}
\usetikzlibrary{patterns,angles,quotes,arrows, shadings, decorations.pathreplacing}
\let\labelindent\relax
\usepackage{enumitem}
\usepackage{algorithm}
\usepackage[noend]{algpseudocode}
\def\thm@space@setup{\thm@preskip=10pt \thm@postskip=0pt}
\makeatother
\newtheoremstyle{mystyle}
  {}
  {}
  {}
  {}
  {\bfseries}
  {.}
  { }
  {}
\newtheorem{remark}{Remark}

\newtheorem{definition}{Definition}
\newtheorem{problem}{Problem}

\DeclareMathOperator{\conv}{conv}

\usepackage{booktabs}
\usepackage{arydshln}

\usepackage{graphicx}
\usepackage{float}
\usepackage[absolute]{textpos}

\begin{document}
\setlength{\abovecaptionskip}{2pt}
\setlength{\belowcaptionskip}{-12pt}

\begin{textblock}{5}(0.5,0.5)
\noindent \textbf{Robotics: Science and Systems 2025} \\
\textbf{Los Angeles, USA, June 21 - 25, 2025}
\end{textblock}

\title{DDAT: Diffusion Policies Enforcing Dynamically Admissible Robot Trajectories}

\author{Author Names Omitted for Anonymous Review. Paper-ID [664]}

\author{\authorblockN{Jean-Baptiste Bouvier, Kanghyun Ryu, Kartik Nagpal, Qiayuan Liao, Koushil Sreenath, Negar Mehr}
\authorblockA{\href{https://iconlab.negarmehr.com/}{ICON Lab} and \href{https://hybrid-robotics.berkeley.edu/}{Hybrid Robotics Lab}, University of California Berkeley}}
\makeatletter
\g@addto@macro\@maketitle{
\begin{figure}[H]
    \setlength{\linewidth}{\textwidth}
    \setlength{\hsize}{\textwidth}
    
    \begin{subfigure}[]{0.49\textwidth}
    \centering
    \begin{tikzpicture}[scale = 1, every node/.style={font=\small}]
    \begin{scope}[scale = 1, xshift=70, yshift=100, rotate=0]
        \def\r{0.1}; 

        \filldraw[left color = green!05, right color = green!30, color=white] (0, -\r) -- (0.9, -0.55) -- (1.1, -0.15) -- (0, \r);
        \filldraw[left color = green!05, right color = green!30, color=white] (1, -0.22-\r) -- (2, -0.7) -- (2, -0.3) -- (1, -0.22+\r);
        \filldraw[left color = green!05, right color = green!30, color=white] (2, -0.4-\r) -- (2.85, -1.05) -- (2.7, -0.55) -- (2, -0.4+\r);
        \filldraw[left color = green!05, right color = green!30, color=white] (2.8, -0.65-\r) -- (3.8, -1.25) -- (4, -0.65) -- (2.8, -0.65+\r);
        \filldraw[left color = green!05, right color = green!30, color=white] (4, -0.75-\r) -- (4.8, -1.25) -- (4.9, -0.85) -- (4, -0.75+\r);

        \filldraw[green!50] (1, -0.35) ellipse (0.3 and 0.2);
        \filldraw[green!50] (2, -0.5) circle (0.2);
        \filldraw[green!50, rotate around={-30:(2.8, -0.8)}] (2.8, -0.8) ellipse (0.3 and 0.2);
        \filldraw[green!50, rotate around={-30:(3.9, -0.95)}] (3.9, -0.95) ellipse (0.2 and 0.3);
        \filldraw[green!50] (4.9, -1.05) circle (0.2);

        \draw[red, ultra thick, dashed] plot [smooth, tension=0.8] coordinates { (0,0) (1, 0.1) (2, 0.) (3, 0.25) (4, 0.) (5, 0.1)};
        \draw[green!50!black, ultra thick, dashed] plot [smooth, tension=0.8] coordinates { (0,0) (1, -0.22) (2, -0.4) (2.8, -0.65) (4, -0.75) (4.95, -0.95)};
        
        \draw[->, very thick, black, dotted] (1, 0.1) -- (1, -0.12);
        \fill[green!60!black] (1, -0.22) circle (\r);
        \draw[->, very thick, black, dotted] (2, 0.) -- (2, -0.3);
        \fill[green!60!black] (2, -0.4) circle (\r);
        \draw[->, very thick, black, dotted] (3, 0.25) -- (2.82, -0.55);
        \fill[green!60!black] (2.8, -0.65) circle (\r);
        \draw[->, very thick, black, dotted] (4, 0.) -- (4, -0.65);
        \fill[green!60!black] (4, -0.75) circle (\r);
        \draw[->, very thick, black, dotted] (5, 0.1) -- (4.97, -0.85);
        \fill[green!60!black] (4.95, -0.95) circle (\r);
        
        \fill[top color=red, bottom color=green!60!black, opacity=0.9]   (0,0) circle (\r);
        \fill[red] (1, 0.1) circle (\r);
        \fill[red] (2, 0.) circle (\r);
        \fill[red] (3, 0.25) circle (\r);
        \fill[red] (4, 0.) circle (\r);
        \fill[red] (5, 0.1) circle (\r);
    \end{scope}

    \node (2) at (1, 4) {$\tau + \varepsilon$};
    
    \node[draw=black, rounded corners=4pt, fill=red!20] (3) at (3.5, 4) {Diffusion model $D_\theta$};
    \draw[->, very thick] (2) to (3);

    \node (4) at (5.5, 4) {\textcolor{red}{$\tilde{\tau}$}};
    \draw[->, very thick] (3) to (4);
    
    \node[align=center, draw=black, rounded corners=4pt, fill=green!20] (5) at (7, 4) {Projection $\mathcal{P}_\sigma$};
    \draw[->, very thick] (4) to (5);
    
    \node (6) at (7.6, 2.2) {\textcolor{green!60!black}{$\tau_p$}};
    \draw[->, very thick] (5) to [out=0,in=0] (6);
    
    \node[align=center] (7) at (3.2, 2.2) {loss $\|\textcolor{green!60!black}{\tau_p} - \tau\|$};
    \draw[->, very thick] (6) to [out=-180,in=0] (7);
    \draw[->, very thick] (7) to [out=180,in=190] (3);
    \node at (7.72, 3.2) {$\mathcal{P}_\sigma$};
        
    \end{tikzpicture}

    \vspace{2mm}
    \begin{tikzpicture}[scale = 1, every node/.style={font=\small}]
        \def\r{0.1}; 

        \node at (-1.8, 0.2) {Complete projection:};

        \filldraw[green!50] (1, -0.35) ellipse (0.3 and 0.2);
        \filldraw[green!50] (2, -0.5) circle (0.2);
        \filldraw[green!50, rotate around={-30:(2.8, -0.8)}] (2.8, -0.8) ellipse (0.3 and 0.2);
        \filldraw[green!50, rotate around={-30:(3.9, -0.95)}] (3.9, -0.95) ellipse (0.2 and 0.3);
        \filldraw[green!50] (4.9, -1.05) circle (0.2);

        \draw[red, ultra thick, dashed] plot [smooth, tension=0.8] coordinates { (0,0) (1, 0.1) (2, 0.) (3, 0.25) (4, 0.) (5, 0.1)};
        \draw[green!50!black, ultra thick, dashed] plot [smooth, tension=0.8] coordinates { (0,0) (1, -0.22) (2, -0.4) (2.8, -0.65) (4, -0.75) (4.95, -0.95)};
        
        \draw[->, very thick, black, dotted] (1, 0.1) -- (1, -0.12);
        \node at (1.2, -0.05) {$e_1$};
        \fill[green!60!black] (1, -0.22) circle (\r);
        \node at (1, -0.42) {\textcolor{green!60!black}{$s_1$}};
        
        \draw[->, very thick, black, dotted] (2, 0.) -- (2, -0.3);
        \node at (2.2, -0.15) {$e_2$};
        \fill[green!60!black] (2, -0.4) circle (\r);
        \node at (2, -0.6) {\textcolor{green!60!black}{$s_2$}};
        
        \draw[->, very thick, black, dotted] (3, 0.25) -- (2.82, -0.55);
        \node at (3.1, -0.2) {$e_3$};
        \fill[green!60!black] (2.8, -0.65) circle (\r);
        \node at (2.8, -0.85) {\textcolor{green!60!black}{$s_3$}};
        
        \draw[->, very thick, black, dotted] (4, 0.) -- (4, -0.65);
        \node at (4.2, -0.35) {$e_4$};
        \fill[green!60!black] (4, -0.75) circle (\r);
        \node at (4, -0.95) {\textcolor{green!60!black}{$s_4$}};
        
        \draw[->, very thick, black, dotted] (5, 0.1) -- (4.97, -0.85);
        \node at (5.2, -0.4) {$e_5$};
        \fill[green!60!black] (4.95, -0.95) circle (\r);
        \node at (4.95, -1.15) {\textcolor{green!60!black}{$s_5$}};
        
        \fill[top color=red, bottom color=green!60!black, opacity=0.9] (0,0) circle (\r);
        \node at (-0.25, 0) {$s_0$};
        \fill[red] (1, 0.1) circle (\r);
        \node at (1, 0.33) {\textcolor{red}{$\tilde{s}_1$}};
        \fill[red] (2, 0.) circle (\r);
        \node at (2, 0.23) {\textcolor{red}{$\tilde{s}_2$}};
        \fill[red] (3, 0.25) circle (\r);
        \node at (3, 0.48) {\textcolor{red}{$\tilde{s}_3$}};
        \fill[red] (4, 0.) circle (\r);
        \node at (4, 0.23) {\textcolor{red}{$\tilde{s}_4$}};
        \fill[red] (5, 0.1) circle (\r);
        \node at (5, 0.33) {\textcolor{red}{$\tilde{s}_5$}};

        \node at (-2, -0.8) {Partial projection:};

        \filldraw[green!50] (-2, -1.85) ellipse (0.3 and 0.2);
        \filldraw[green!50] (-1, -2) circle (0.2);
        \filldraw[green!50, rotate around={-30:(-0.2, -2.3)}] (-0.2, -2.3) ellipse (0.3 and 0.2);
        \filldraw[green!50] (1.9, -1.9) circle (0.2);

        \draw[red, ultra thick, dashed] plot [smooth, tension=0.8] coordinates { (-3, -1.5) (-2, -1.4) (-1, -1.5) (0, -1.25) (1, -1.5) (2, -1.4)};
        \draw[green!50!black, ultra thick, dashed] plot [smooth, tension=0.8] coordinates { (-3, -1.5) (-2, -1.72) (-1, -1.9) (-0.2, -2.15)};
        \draw[green!50!black, ultra thick, dashed, draw opacity=0.2] plot [smooth, tension=0.8] coordinates { (-0.2, -2.15) (1, -1.5)};
        \draw[green!50!black, ultra thick, dashed] plot [smooth, tension=0.8] coordinates { (1, -1.5) (1.95, -1.8)};
        
        \draw[->, very thick, black, dotted] (-2, -1.4) -- (-2, -1.62);
        \node at (-1.8, -1.55) {$e_1$};
        \fill[green!60!black] (-2, -1.72) circle (\r);
        \node at (-2, -1.92) {\textcolor{green!60!black}{$s_1$}};
        
        \draw[->, very thick, black, dotted] (-1, -1.5) -- (-1, -1.8);
        \node at (-0.8, -1.65) {$e_2$};
        \fill[green!60!black] (-1, -1.9) circle (\r);
        \node at (-1, -2.1) {\textcolor{green!60!black}{$s_2$}};
        
        \draw[->, very thick, black, dotted] (0, -1.25) -- (-0.18, -2.05);
        \node at (0.1, -1.7) {$e_3$};
        \fill[green!60!black] (-0.2, -2.15) circle (\r);
        \node at (-0.2, -2.35) {\textcolor{green!60!black}{$s_3$}};
        
        \draw[->, very thick, black, dotted] (2, -1.4) -- (1.97, -1.7);
        \node at (2.2, -1.6) {$e_5$};
        \fill[green!60!black] (1.95, -1.8) circle (\r);
        \node at (1.95, -2.) {\textcolor{green!60!black}{$s_5$}};
        
        \fill[top color=red, bottom color=green!60!black, opacity=0.9] (-3,-1.5) circle (\r);
        \node at (-3., -1.3) {$s_0$};
        \fill[red] (-2, -1.4) circle (\r);
        \node at (-2, 0.33-1.5) {\textcolor{red}{$\tilde{s}_1$}};
        \fill[red] (-1, -1.5) circle (\r);
        \node at (-1, 0.23-1.5) {\textcolor{red}{$\tilde{s}_2$}};
        \fill[red] (0, -1.25) circle (\r);
        \node at (0, -1.02) {\textcolor{red}{$\tilde{s}_3$}};
        \fill[top color=red, bottom color=green!60!black, opacity=0.9] (1, -1.5) circle (\r);
        \node at (1, 0.23-1.5) {$s_4$};
        \fill[red] (2, -1.4) circle (\r);
        \node at (2, 0.33-1.5) {\textcolor{red}{$\tilde{s}_5$}};
    \end{tikzpicture}
 
    \end{subfigure} \hfill
    \begin{subfigure}[]{0.221\textwidth}
        \centering
        \includegraphics[scale=0.211]{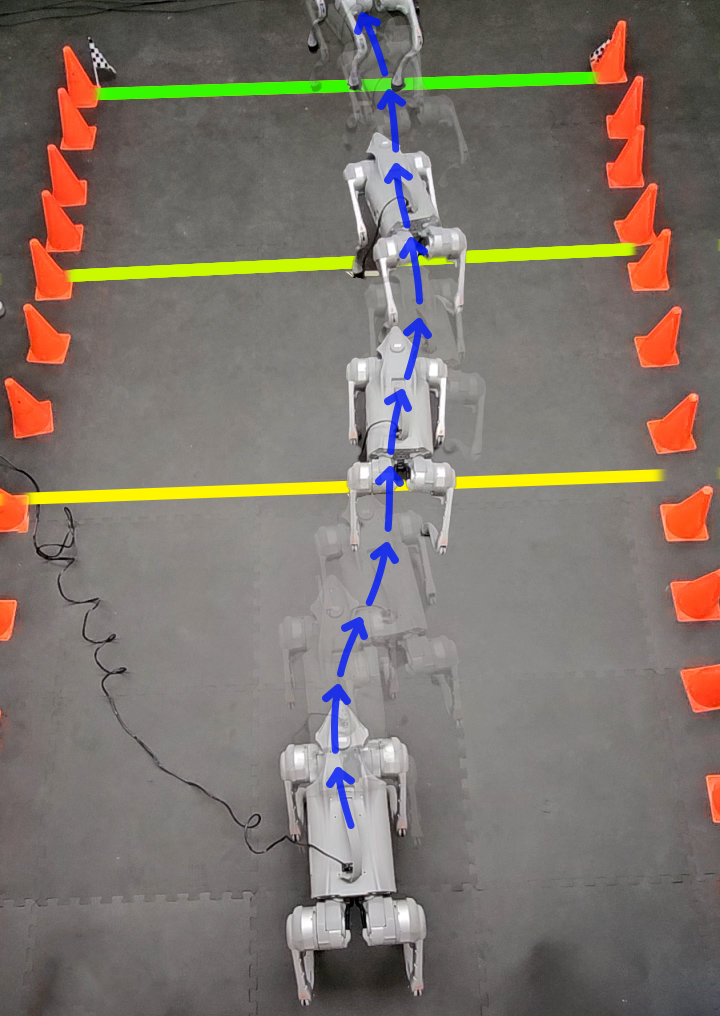}
    \end{subfigure}
    \begin{subfigure}[]{0.27\textwidth}
        \centering
        \includegraphics[scale=0.28]{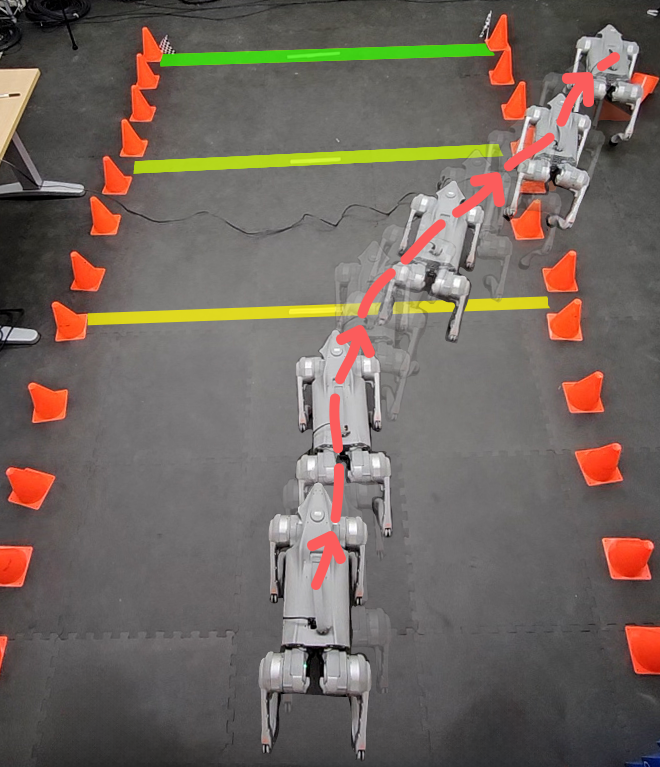}
    \end{subfigure}
    \caption{(top left) Overview of \emph{Diffusion policies for Dynamically Admissible Trajectories} (DDAT). Diffusion model $D_\theta$ is trained to predict a trajectory \textcolor{red}{$\tilde{\tau}$} given a trajectory $\tau$ from the training dataset corrupted by noise $\varepsilon$. If the noise level $\sigma$ of signal $\varepsilon$ is sufficiently small, $\mathcal{P}_\sigma$ projects \textcolor{red}{$\tilde{\tau}$} to the dynamically admissible trajectory \textcolor{green!50!black}{$\tau_p$}. The loss $\| \textcolor{green!50!black}{\tau_p} - \tau\|$ is used to update $D_\theta$. 
    (bottom left) To smoothly incorporate projections into the training and inference of our diffusion models we use partial projections of infeasible trajectories. The total projection error $\sum e_t$ is significantly reduced by not projecting \textcolor{red}{$\tilde{s}_4$} on the reachable set of \textcolor{green!60!black}{$s_3$}.
    (right) Demonstration of DDAT on the Unitree GO2. A vanilla diffusion policy fails at walking through the cones in open-loop. By accounting for the quadruped's dynamics our open-loop diffusion policy succeeds in following the corridor.}
    \label{fig: DDAT scheme}
    \end{figure}
    \vspace{-3mm}
}
\maketitle

\begin{abstract}
    Diffusion models excel at creating images and videos thanks to their multimodal generative capabilities. These same capabilities have made diffusion models increasingly popular in robotics research, where they are extensively used for generating robot motion.
    However, the stochastic nature of diffusion models is fundamentally at odds with the precise dynamical equations describing the feasible motion of robots. Hence, generating dynamically admissible robot trajectories is a challenge for diffusion models.
    To alleviate this issue, we introduce \emph{DDAT: \underline{D}iffusion policies for \underline{D}ynamically \underline{A}dmissible \underline{T}rajectories} to generate admissible trajectories of black-box robotic systems using diffusion models.
    To generate such trajectories, our diffusion policies project their predictions onto a dynamically admissible manifold during both training and inference to align the objective of the denoiser neural network with the dynamical admissibility constraint. Due to the auto-regressive nature of such projections as well as the black-box nature of robot dynamics, trajectory projections are challenging.  
    We thus enforce admissibility by iteratively sampling a polytopic under-approximation of the reachable set of a state onto which we project its predicted successor, before iterating this process with the projected successor.
    By producing accurate trajectories, this projection eliminates the need for diffusion models to continually replan, enabling \emph{one-shot long-horizon trajectory planning}.
    We demonstrate that our proposed framework generates higher quality dynamically admissible robot trajectories through extensive simulations on a quadcopter and various MuJoCo environments, along with real-world experiments on a Unitree GO1 and GO2. Code available on our website: \href{https://iconlab.negarmehr.com/DDAT/}{https://iconlab.negarmehr.com/DDAT/}.
\end{abstract}

\IEEEpeerreviewmaketitle

\section{Introduction}

Diffusion models have achieved state-of-the-art performance in image and video generation thanks to their multimodal generative capabilities~\citep{DDPM, SMLD, song2020score}. In robotics planning and control tasks these diffusion models have shown promise thanks to their ability to generate entire trajectories at once, avoiding compounding errors and improving long-term performance~\citep{Visuomotor, Diffuser}.
These diffusion planners generate trajectories by iteratively reducing the noise of an initial random sequence until obtaining a trajectory without noise~\citep{Diffuser}. However, this stochastic denoising process is fundamentally at odds with the precise equations describing the dynamically admissible motion of robots. 
Indeed, the deterministic equations of motion of a robot only consider a transition from state $s_t$ to $s_{t+1}$ to be \emph{admissible} if $s_{t+1}$ is in the reachable set of $s_t$. This reachable set is typically a bounded manifold of dimension much lower than the state space due to the robot's limited actuation~\citep{underactuated}.  
Learning to sample exclusively from such reachable sets is extremely challenging due to their small size and their dependence on the state $s_t$. Hence, diffusion models typically generate inadmissible sequences of states for underactuated robots, which has mostly limited the deployment of diffusion planners to fully-actuated systems like Maze2D~\citep{d4rl} and robotic arms~\citep{Trajectory, Diffuser, AdaptDiffuser, lee2024refining, romer2024diffusion, saha2024edmp, sun2024conformal, Replanning}.

The typical approach to deploy diffusion planning on underactuated robots is to regenerate an entire trajectory every few timesteps as the state actually reached by the robot differs from the one predicted with the inadmissible trajectory generated by the diffusion model~\citep{Trajectory, DiffuserLite, DiffuseLoco, CGD, Replanning}.
However, this approach is computationally demanding and wasteful.
More promising is to project the predicted trajectory onto the dynamically admissible manifold after inference~\citep{giannone2024aligning, CGD, beat_GAN, power2023sampling, romer2024diffusion, D-MPC}. 
However, this post-inference projection can cause divergence when the trajectory is too far from being admissible~\citep{PhysDiff}. For the diffusion policy to account for this projection and correct its prediction, works~\citep{Projected, giannone2024aligning, romer2024safe, PhysDiff} have instead used projections throughout inference.

To address the dynamic feasibility issue in diffusion planning from offline data, we introduce \textbf{Diffusion policies for Dynamically Admissible Trajectories (DDAT)}. These models use a chronological trajectory projector capable of transforming a generated sequence of states into an admissible trajectory. On the contrary to work~\citep{CGD}, and inspired by \citep{Projected, giannone2024aligning, PhysDiff}, the projections in DDAT occurs throughout and after inference instead of only after inference. Differing from works~\citep{Projected, giannone2024aligning, PhysDiff}, these projections are also implemented during training of the diffusion models to align their training objectives with their inference phase, as suggested in work~\citep{bastek2024physics}. Furthermore, works~\citep{bastek2024physics, Projected, giannone2024aligning, PhysDiff} only concern themselves with a single constraint, which is significantly less challenging to enforce than dynamical admissibility due to its autoregressive nature. 
Another advantage of generating feasible trajectories is that we do not need to sample and denoise a large batch of trajectories hoping to find an admissible one, which provides significant computational gain compared to methods like~\citep{carvalho2023motion}. Only works~\citep{gadginmath2025dynamics, romer2024diffusion} studied dynamical feasibility in diffusion planning, but both works focus solely on linear systems.

Our main contributions in this work are as follows.
\begin{enumerate}
    \item We generate dynamically admissible trajectories with diffusion models for black-box systems.
    \item We design an algorithm to train diffusion models to respect dynamical admissibility constraints.
    \item We compare various projections and diffusion model architectures to enforce dynamic admissibility.
\end{enumerate}

The remainder of this work is organized as follows. 
In Section~\ref{sec: related works}, we provide a survey of related works.
In Section~\ref{sec: framework}, we introduce our problem formulation along with our framework.
In Section~\ref{sec: projections}, we define several projectors to enforce the dynamical admissibility of trajectories.
In Section~\ref{sec: DDAT}, we incorporate these projectors into diffusion models.
In Section~\ref{sec: implementation}, we implement our proposed Diffusion policies for Dynamically Admissible Trajectories.
We present our simulation results on a quadcopter, the Gym Hopper, Walker and HalfCheetah, and the Unitree GO1 and GO2 quadrupeds in Section~\ref{sec: simulations}, while Section~\ref{sec: hardware} demonstrates the hardware implementation of DDAT on a Unitree GO1 and a GO2.

\textit{Notation:}
The positive integer interval from $a \in \mathbb{N}$ to $b \in \mathbb{N}$ inclusive is denoted by $[\![a, b]\!]$. The convex hull of a set of points $p_1, ..., p_m \in \mathbb{R}^n$ is denoted as $\conv\big(p_1, ..., p_m\big) := \big\{ \sum_{i=1}^m \lambda_i p_i : \lambda_i \in [0,1],\ \sum_{i=1}^m \lambda_i = 1\big\}$. A value $a$ sampled uniformly from a set $\mathcal{A}$ is denoted as $a \sim \mathcal{U}\big(\mathcal{A}\big)$. A Bernoulli variable $b$ taking value $1$ with probability $p$ and $0$ otherwise is denoted by $b \sim \mathcal{B}(p)$.

\section{Related Works}\label{sec: related works}

In this section, we review the literature relevant to our area of interest, namely constraint enforcement on diffusion models and diffusion planning.

\subsection{Enforcing constraints on diffusion models}

\subsubsection{Soft constraints}

Most diffusion works employ \emph{soft constraints} to encourage satisfaction of a given constraint, either with classifier guidance~\citep{Trajectory, RyneBeeson, beat_GAN}, classifier-free guidance~\citep{power2023sampling}, or simply using additional loss terms~\citep{bastek2024physics, carvalho2023motion, giannone2024aligning, sun2024conformal, Diff-RNTraj}, either only during inference~\citep{carvalho2023motion, giannone2024aligning} or during both training and inference~\citep{bastek2024physics, Trajectory, RyneBeeson, sun2024conformal, Diff-RNTraj}.
However, these soft constraints cannot guarantee satisfaction for all the samples generated by the diffusion process. Thus, most of these works need to sample large batches of trajectories among which they try to select the prediction closest to satisfying the constraint~\citep{Trajectory, carvalho2023motion}.

\subsubsection{Hard constraints}

Contrasting the works presented above, we are interested in hard constraints to guarantee their satisfaction. 
The first hard constraints imposed on diffusion models were $[0, 255]$ pixel cutoffs, which led to saturated images~\citep{Reflected}. To mitigate this quality loss,~\citep{Reflected} proposed to replace cutoffs with reflection on the constraints boundaries. In addition,~\citep{constraints} extended this reflection to manifolds and introduced log-barrier as a constraint enforcer on manifolds.
When saturation is not an issue, projection remains the most widely adopted approach to enforce constraints~\citep{Projected, CGD, romer2024safe, romer2024diffusion, PhysDiff}.
While a single projection at the end of inference is sufficient to enforce a constraint~\citep{CGD}, works~\citep{Projected, PhysDiff} realized that several projection steps interleaved with the denoising process yield higher quality constraint-satisfying samples.
Bringing these advances to diffusion planning,~\citep{romer2024safe} enforces state and action bounds by projecting partially denoised trajectories during inference. 
Most of these works also agree that constraints should mostly be enforced at low noise levels where the sampled predictions are not just random sequences~\citep{bastek2024physics, RyneBeeson, PhysDiff}. However, none of these works enforces dynamic feasibility constraints, whose autoregressive nature renders admissible diffusion planning significantly more challenging.

\subsection{Diffusion planning}\label{subsec: litt planning}

We will now review the literature on diffusion planning and discuss how they handle dynamical feasibility.

\subsubsection{Dynamic feasibility enforced by planning on actions}

The easiest solution to generate dynamically admissible trajectories is to let the diffusion model directly generate sequences of actions since the resulting sequence of states will be admissible by definition~\citep{Visuomotor, RyneBeeson, Diffusion_QLearning, zhong2023guided}. The problem with this approach is the high variability and lack of smoothness of sequences of actions, rendering them much more challenging to predict than sequences of states and hence leading to lower quality plans~\citep{DecisionDiffuser}. This observation led works~\citep{BESO, NoMaD} to use transformers and work~\citep{D-MPC} to use diffusion models to learn the temporal relations between states and actions, and a diffusion model to predict actions. These added models create an unnecessary computational overhead only to mitigate the difficulty of planning over actions.

\subsubsection{Diffusion planning on states}

The most straightforward approach to diffusion planning is to let the diffusion model predict the sequence of states directly. This sequence of states can be a reward maximizing trajectory~\citep{DecisionDiffuser} or predicted states of other agents~\citep{gu2022stochastic}. However, these state trajectories are generated through a stochastic process and thus have no guarantees of admissibility. This observation prompted replanning approaches, which generate an entire new trajectory every few timesteps as the state actually reached by the robot differs from the one predicted with the potentially inadmissible trajectory generated by the diffusion model~\citep{Replanning}. Due to the slow nature of replanning with diffusion models,~\citep{DiffuserLite, DiffuseLoco} focus on accelerating diffusion inference for real-time planning. Other works are using restoration gaps~\citep{lee2024refining} and constraint gradients~\citep{saha2024edmp} to guide their diffusion models, but none of these works guarantee satisfaction of the robots dynamics.

Most closely related to this work is the emerging literature enforcing constraints in diffusion planning with quadratic programming~\citep{CGD}, control barrier functions~\citep{SafeDiffuser}, or projections~\citep{romer2024safe}. Only works~\citep{gadginmath2025dynamics, romer2024diffusion} consider dynamic constraints, but only test their approaches on linear fully-actuated systems.

\subsubsection{Diffusion planning on states and actions}

Joint prediction of state and actions can enhance planning quality by improving temporal coherence of planned trajectories~\citep{Diffuser}. However, Diffuser~\citep{Diffuser} only learns this coherence from its training data and does not enforce it with a dynamics model, leading to infeasible trajectories. AdaptDiffuser~\citep{AdaptDiffuser} uses neural network-based inverse dynamics to iteratively revise the state sequence and predicted reward. However, this learned inverse model may differ from the true inverse dynamics and allow infeasible trajectories. As a result,~\citep{RyneBeeson, power2023sampling} use diffusion models only to generate initial guess for trajectory optimization solvers capable of enforcing dynamic feasibility. Thus, none of the diffusion planning literature formally addresses the dynamic admissibility problem.

\section{Framework}\label{sec: framework}

Let us now introduce our framework. We consider a robot of state $s_t$ at time $t$ with \emph{black-box} deterministic discrete-time dynamics of form
\begin{equation}\label{eq: nonlinear dynamics}
    s_{t + 1} = f\big( s_t, a_t \big), \quad a_t \in \mathcal{A}, \quad s_0 \sim \mathcal{U}(\mathcal{S}_0),
\end{equation}
where action $a_t$ belongs to the admissible action set $\mathcal{A} \subset \mathbb{R}^m$ and $\mathcal{S}_0$ is the set of initial states in the state space $\mathcal{S} \subseteq \mathbb{R}^n$. We emphasize that dynamics~\eqref{eq: nonlinear dynamics} are a \emph{black-box}, i.e. the equations describing $f$ are not available, but the user can obtain $s_{t+1}$ given $s_t$ and $a_t$. This setting corresponds to a numerical simulator which is typically available for any robot.
We now want to use diffusion policies to generate trajectories of robot~\eqref{eq: nonlinear dynamics}.

\subsection{Diffusion planning}\label{subsec: planning}

Diffusion planning refers to using diffusion models~\citep{DDPM, SMLD, song2020score} to generate trajectories of a given system. 
Due to the black-box nature of dynamics~\eqref{eq: nonlinear dynamics}, this imitation learning approach requires a training dataset $\mathcal{D}$ of example trajectories representative of a desired data distribution $p_{data}$.
A trajectory of length $H+1$ is a sequence of states $\tau := \big\{s_0, s_1, ..., s_{H}\big\} \in \mathcal{S}^{H+1}$.
The objective is to learn how to sample trajectories from $p_{data}$. To do so, a forward diffusion process starts by corrupting dataset $\mathcal{D}$ with Gaussian noise $\varepsilon \sim \mathcal{N}(0, \sigma^2 I)$ parametrized by $\sigma$ sampled as $\ln(\sigma) \sim \mathcal{N}(p_m, p_s^2)$ with parameters from \citep{karras2022elucidating} defined in Appendix~\ref{apx: diffusion}. Diffusion planning can be viewed as learning to reverse this process. More specifically, we iteratively reduce the noise level of a randomly sampled sequence until obtaining a trajectory belonging to $p_{data}$, this process is called \emph{denoising}. We train a neural network $D_\theta$ to generate denoised trajectories from corrupted ones $\tau + \varepsilon$ by minimizing
\begin{equation*}
    \underset{\theta}{\min} \underset{\ln(\sigma) \sim \mathcal{N}(p_m, p_s^2)}{\mathbb{E}} \ \underset{\varepsilon \sim \mathcal{N}(0, \sigma^2 I)}{\mathbb{E}} \ \underset{\tau \sim \mathcal{U}(\mathcal{D})}{\mathbb{E}} \big\| D_\theta( \tau + \varepsilon; \sigma) - \tau \big\|^2.
\end{equation*}
To generate trajectories from the desired distribution $p_{data}$ we
implement a first-order deterministic sampler relying on $D_\theta$ and detailed in Appendix~\ref{apx: diffusion} simplified from \citep{karras2022elucidating}. This sampler starts from a random sequence $\tau_0 \sim \mathcal{N}(0, \sigma_{0}^2 I)$ and iteratively reduces its noise level from $\sigma_i$ to $\sigma_{i+1}$ with $\tau_{i+1} = S_\theta(\tau_i; \sigma_i, \sigma_{i+1})$ where $\sigma_0 >\!\!> 1 $ and $\sigma_N = 0$. The derivation of denoising operator
\begin{equation}\label{eq: denoising operator}
    S_\theta(\tau_i; \sigma_i, \sigma_{i+1}) := \frac{\sigma_{i+1}}{\sigma_i}\tau_i + \Big( 1 - \frac{\sigma_{i+1}}{\sigma_i} \Big) D_\theta(\tau_i, \sigma_i)
\end{equation}
is detailed in Appendix~\ref{apx: diffusion}.
After $N$ iterations we eventually obtain a trajectory $\tau_N \sim p_{data}$.

\subsection{Problem formulation}

\begin{definition}\label{def: admissibility}
    A trajectory $\{s_0, ..., s_{H}\}$ is \emph{admissible} by dynamics~\eqref{eq: nonlinear dynamics} if and only if there exist a sequence of actions $\big\{a_0, ..., a_{H-1}\big\} \in \mathcal{A}^H$ such that $s_{t+1} = f(s_t, a_t)$ for all $t \in [\![0, H-1]\!]$.
\end{definition}
Equivalently, admissibility requires each state $s_{t+1}$ to belong to the reachable set of its predecessor $s_t$.
\begin{definition}\label{def: reachable set}
    The \emph{reachable set} $\mathcal{R}(s_t)$ of a state $s_t$ is the set of all feasible next states:
    \begin{equation}\label{eq: reachable set}
        \mathcal{R}(s_t) := \big\{ f(s_t, a) : \ \emph{\text{for all}}\ a \in \mathcal{A} \big\}.
    \end{equation}
\end{definition}

Throughout this work, we will emphasize the difference between \emph{sequences of states}, where iterated states are not necessarily reachable from their predecessor, and \emph{admissible trajectories} satisfying exactly and recursively dynamics~\eqref{eq: nonlinear dynamics}.
This distinction leads us to our problem of interest.

\begin{problem}\label{prob: admissible diffusion planning}
    How to generate dynamically admissible trajectories for a black-box model using diffusion policies?
\end{problem}

While diffusion policies excel at generating sequences of states~\citep{DiffuserLite, Diffuser, AdaptDiffuser}, the difficulty of Problem~\ref{prob: admissible diffusion planning} resides in guaranteeing their dynamical admissibility, which is challenging for two reasons. 
The first difficulty comes from the chronological nature of the required projections of each $s_{t+1}$ onto the reachable set $\mathcal{R}(s_t)$, which are computationally expensive and cannot be parallelized.
The second challenge resides in the black-box nature of dynamics~\eqref{eq: nonlinear dynamics}, hence preventing a closed-form expression of the reachable sets $\mathcal{R}(s_t)$ of~\eqref{eq: reachable set}. Even with known nonlinear dynamics, approximating reachable sets is still a challenging and active research topic~\citep{convexHull}. We will discuss how to address these challenges in the next section.

\section{Making Trajectories Admissible}\label{sec: projections}

In this section we propose several projections schemes to make the trajectories generated by our diffusion models become dynamically admissible.
Each one of these approaches have their own pros and cons in terms of computation, admissibility guarantees, and whether they require action predictions along with the state predictions.

\subsection{Reachable sets underapproximation}

To ensure dynamic feasibility of a sequence of states $\{s_0, \tilde{s}_1, ..., \tilde{s}_H\}$, we need to chronologically project each $\tilde{s}_{t+1}$ onto the reachable set of its predecessor $\mathcal{R}(s_t)$. However, the black-box nature of dynamics~\eqref{eq: nonlinear dynamics} prevents any closed-form expression of the reachable set~\eqref{eq: reachable set}. To render this problem tractable, we will assume the convexity of the reachable set and underapproximate it with a polytope. This assumption holds for sufficiently smooth nonlinear systems and small timesteps~\citep{azhmyakov2007convexity} but fails for robots with instantaneous contact forces. Fortunately, we will see in our experiments of Section~\ref{sec: simulations} that the nonconvexity of the reachable set has but a small impact on our approach.

We assume to know a polytopic underapproximation of the admissible action set $\mathcal{A}$ with vertices $v_1, ..., v_q \in \mathcal{A}$ such that $\conv\big(v_1, ..., v_q\big) \subseteq \mathcal{A}$. This assumption is typically verified as $\mathcal{A}$ is often a known hyperrectangle of dimension $m$ delimited by $q = 2^m$ vertices~\cite{Gym}.
Then, given a state $s_t$, we underapproximate its reachable set with
\begin{equation}\label{eq: reachable set approx}
    \mathcal{C}(s_t) := \conv\big( f(s_t,v_1), ..., f(s_t,v_q) \big),
\end{equation}
which is the convex hull of the successors of $s_t$ by black-box dynamics $f$ of~\eqref{eq: nonlinear dynamics} and actions $v_i$. We now need to iteratively project each predicted next state $\tilde{s}_{t+1}$ onto this tractable approximation of the reachable set of $s_t$ starting from $s_0$ with projector
\begin{equation}\label{eq: projection}
    \mathcal{P}\big(\tilde{s}_{t+1}, \mathcal{C}(s_t)\big) := \underset{c\, \in\, \mathcal{C}(s_t)}{\arg\min} \|\tilde{s}_{t+1} - c\|.
\end{equation}

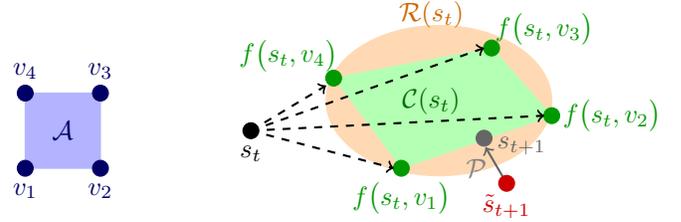
\begin{figure}[t]
    \centering
    \begin{tikzpicture}[scale = 1]
        \def\r{3pt}; 
    
        \filldraw[blue!30] (-3, -0.5) rectangle (-2, 0.5);
        \node at (-2.5, 0) {\textcolor{blue!40!black}{$\mathcal{A}$}};
        \filldraw[blue!40!black] (-3, -0.5) circle (\r);
        \node at (-3, -0.8) {\textcolor{blue!40!black}{$v_1$}};
        \filldraw[blue!40!black] (-3, 0.5) circle (\r);
        \node at (-3, 0.8) {\textcolor{blue!40!black}{$v_4$}};
        \filldraw[blue!40!black] (-2, -0.5) circle (\r);
        \node at (-2, -0.8) {\textcolor{blue!40!black}{$v_2$}};
        \filldraw[blue!40!black] (-2, 0.5) circle (\r);
        \node at (-2, 0.8) {\textcolor{blue!40!black}{$v_3$}};

        \filldraw[orange!30] (2.5, 0.4) ellipse (1.5 and 1);
        \node at (2.4, 1.55) {\textcolor{orange!80!black}{$\mathcal{R}(s_t)$}};
        \filldraw[green!30] (2, -0.5) -- (4, 0.2) -- (3.2, 1.1) -- (1.1, 0.7);
        \node at (2.4, 0.4) {\textcolor{green!40!black}{$\mathcal{C}(s_t)$}};        

        \filldraw[green!60!black] (2, -0.5) circle (\r);
        \node at (2, -0.9) {\textcolor{green!60!black}{$f\big(s_t, v_1\big)$}};
        \draw[->, thick, dashed] (0, 0) -- (1.9, -0.5);

        \filldraw[green!60!black] (4, 0.2) circle (\r);
        \node at (4.8, 0.2) {\textcolor{green!60!black}{$f\big(s_t, v_2\big)$}};
        \draw[->, thick, dashed] (0, 0) -- (3.9, 0.2);

        \filldraw[green!60!black] (3.2, 1.1) circle (\r);
        \node at (3.9, 1.35) {\textcolor{green!60!black}{$f\big(s_t, v_3\big)$}};
        \draw[->, thick, dashed] (0, 0) -- (3.1, 1.1);

        \filldraw[green!60!black] (1.1, 0.7) circle (\r);
        \node at (0.5, 1.0) {\textcolor{green!60!black}{$f\big(s_t, v_4\big)$}};
        \draw[->, thick, dashed] (0, 0) -- (1, 0.6);

        \filldraw[black] (0, 0) circle (\r);
        \node at (0, -0.3) {$s_t$};

        \draw[->, thick, black!60] (3.4, -0.7) -- (3.12, -0.22);
        \node at (3.0, -0.5) {\textcolor{black!50}{$\mathcal{P}$}};
        \filldraw[red!80!black] (3.4, -0.7) circle (\r); 
        \node at (3.4, -1.) {\textcolor{red!80!black}{$\tilde{s}_{t+1}$}};
        \filldraw[black!60] (3.1, -0.1) circle (\r);
        \node at (3.6, -0.2) {\textcolor{black!60}{$s_{t+1}$}};
        
    \end{tikzpicture}
    \caption{Illustration of trajectory projection~\eqref{eq: projection}. The extremal actions \textcolor{blue!40!black}{$v_1, v_2, v_3, v_4$} of the admissible action set \textcolor{blue!40!black}{$\mathcal{A}$} are applied to $s_t$ with dynamics $f$ to get extremal next states \textcolor{green!60!black}{$f\big(s_t, v_1\big), f\big(s_t, v_2\big), f\big(s_t, v_3\big), f\big(s_t, v_4\big)$}. Their convex hull generates a polytope \textcolor{green!60!black}{$\mathcal{C}(s_t)$} underapproximating the actual reachable set \textcolor{orange!80!black}{$\mathcal{R}(s_t)$}. The predicted next state \textcolor{red!50!black}{$\tilde{s}_{t+1}$} is then projected by \textcolor{black!60}{$\mathcal{P}$} onto \textcolor{green!60!black}{$\mathcal{C}(s_t)$} to obtain admissible next state \textcolor{black!60}{$s_{t+1}$}.}
    \label{fig: proj}
    \vspace{4mm}
\end{figure}

We illustrate a projection step in Fig.~\ref{fig: proj} and summarize the whole trajectory projection in Algorithm~\ref{alg: proj}.

\begin{remark}\label{rmk: inadmissible proj}
    The projected trajectory $\tau$ computed by Algorithm~\ref{alg: proj} is not necessarily admissible as discontinuities of dynamics~\eqref{eq: nonlinear dynamics} can make the actual reachable set $\mathcal{R}(s_t)$ nonconvex. Then, $\mathcal{C}(s_t)$ might not be a subset of $\mathcal{R}(s_t)$ and the projected next state can be infeasible if $\mathcal{P}\big(\tilde{s}_{t+1}, \mathcal{C}(s_t)\big) \notin \mathcal{R}(s_t)$.
\end{remark}

\begin{algorithm}[htb!]
\caption{State Trajectory Projection}\label{alg: proj}
\begin{algorithmic}[1]
\Require predicted trajectory $\tilde{\tau} = \big\{s_0, \tilde{s}_1,... \tilde{s}_H\big\} \in \mathcal{S}^{H+1}$. 
\vspace{2mm}
\State $\tau = \big\{ s_0 \big\}$ \Comment{initialize the projected trajectory}
\For{$t=0$ to $H-1$}
\State $\mathcal{C}(s_t) = \conv\big( f(s_t,v_1), ..., f(s_t,v_m) \big)$
\State $s_{t+1} = \mathcal{P}\big(\tilde{s}_{t+1}, \mathcal{C}(s_t)\big)$ \Comment{next state projection}
\State $\tau \leftarrow \tau \cup \{s_{t+1}\}$ \Comment{append to $\tau$}
\EndFor
\State \textbf{return} $\tau$
\end{algorithmic}
\end{algorithm}

\vspace{-2mm}
\subsection{Reference projection of state trajectories}\label{subsec: reference projs}

\begin{figure}[b!]
    \vspace{-2mm}
    \centering
    \includegraphics[scale=0.5]{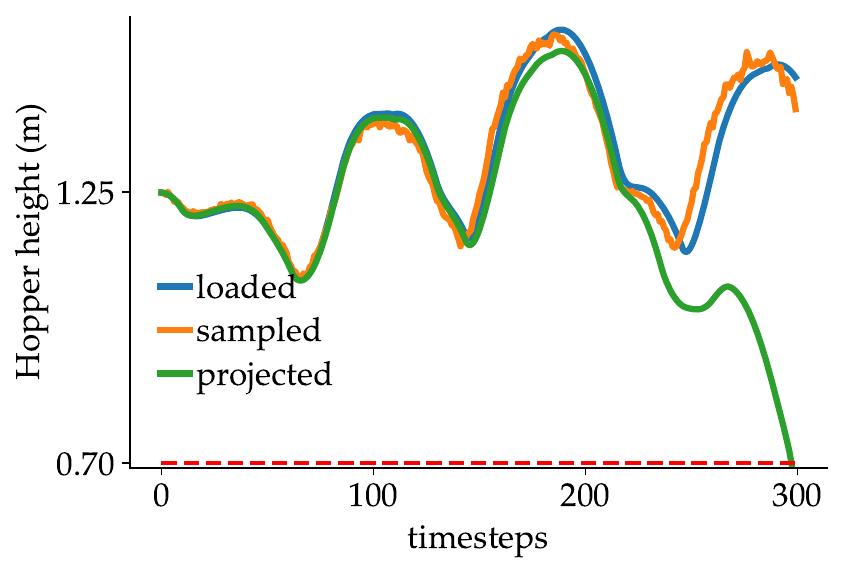}
    \vspace{-2mm}
    \caption{Diverging projection of the Hopper height. Our diffusion model samples the \textcolor{orange!90!black}{\textbf{orange}} trajectory given the initial state of the \textcolor{blue!70}{\textbf{blue}} trajectory, which is loaded from our dataset. The projection of \textcolor{orange!90!black}{\textbf{orange}} with Algorithm~\ref{alg: proj} yields the diverging but admissible \textcolor{green!60!black}{\textbf{green}} trajectory.
    However, \textcolor{green!60!black}{\textbf{green}} is not the admissible trajectory closest to \textcolor{orange!90!black}{\textbf{orange}} since \textcolor{blue!70}{\textbf{blue}} is admissible and much closer to \textcolor{orange!90!black}{\textbf{orange}} than \textcolor{green!60!black}{\textbf{green}}. }
    \label{fig: hopper proj}
\end{figure}

While Algorithm~\ref{alg: proj} makes a trajectory admissible, its iterative nature leads to compounding errors and diverging trajectories. For instance, Fig.~\ref{fig: hopper proj} shows that Algorithm~\ref{alg: proj} cannot project a Hopper trajectory~\cite{Gym} onto the closest admissible trajectory and instead leads to divergence.
Indeed, the greedy nature of Algorithm~\ref{alg: proj} minimizes the distance at the current step without accounting for the divergence it may cause later on. A dynamic programming approach could anticipate this issue but at a prohibitive computational cost.

To compensate for the compounding errors of Algorithm~\ref{alg: proj}, we instead propose to guide projection $\mathcal{P}$ of~\eqref{eq: projection} towards an admissible next state better aligned with the desired trajectory using a reference trajectory. More specifically, we argue that the projection of $\tilde{s}_{t+1}$ should also minimize the distance to a reference next state $s^\text{ref}_{t+1}$ as follows:
\begin{equation}\label{eq: ref proj}
    \mathcal{P}^\text{ref}\big(\tilde{s}_{t+1}, \mathcal{C}(s_t), s_{t+1}^\text{ref} \big) := \underset{c\, \in\, \mathcal{C}(s_t)}{\arg \min} \big\{ \|\tilde{s}_{t+1} - c\| + \lambda \|s_{t+1}^\text{ref} - c\| \big\},
\end{equation}
where $\lambda > 0$ is a trade-off coefficient between the projection and the proximity to the reference. with $\mathcal{P}^\text{ref} = \mathcal{P}$ when $\lambda = 0$.
Replacing $\mathcal{P}$ with $\mathcal{P}^\text{ref}$ in line 3 of Algorithm~\ref{alg: proj} yields the reference trajectory projection Algorithm~\ref{alg: ref proj}. 

\begin{algorithm}[htb!]
\caption{Reference Trajectory Projection}\label{alg: ref proj}
\begin{algorithmic}[1]
\Require predicted trajectory $\tilde{\tau} = \big\{s_0, \tilde{s}_1,... \tilde{s}_H\big\} \in \mathcal{S}^{H+1}$, reference trajectory $\tau^\text{ref} = \big\{s_0, s_1^\text{ref},... s_H^\text{ref} \big\} \in \mathcal{S}^{H+1}$.
\vspace{2mm}
\State $\tau = \big\{ s_0 \big\}$ \Comment{initialize the projected trajectory}
\For{$t=0$ to $H-1$}
\State $\mathcal{C}(s_t) = \conv\big( f(s_t,v_1), ..., f(s_t,v_m) \big)$
\State $s_{t+1} = \mathcal{P}^\text{ref}\big(\tilde{s}_{t+1}, \mathcal{C}(s_t), s^\text{ref}_{t+1} \big)$ \Comment{next state projection}
\State $\tau \leftarrow \tau \cup \{s_{t+1}\}$ \Comment{append to $\tau$}
\EndFor
\State \textbf{return} $\tau$
\end{algorithmic}
\end{algorithm}

\subsection{Using action predictions for better state projections}

While Algorithms~\ref{alg: proj} and \ref{alg: ref proj} perform relatively well in low-dimensional state spaces such as the Hopper of Gymnasium~\citep{Gym}, these projections are not sufficiently precise to recreate admissible trajectories $\tau$ of the Walker2D and HalfCheetah as $\big\| \mathcal{P}(\tau) - \tau \big\| >\!\!> 0$. This accuracy degradation is most likely caused by the non-convexity of the reachable sets following Remark~\ref{rmk: inadmissible proj} and the increased action space dimension which enlarges set~\eqref{eq: reachable set approx} and keeps the projection far from the reference.

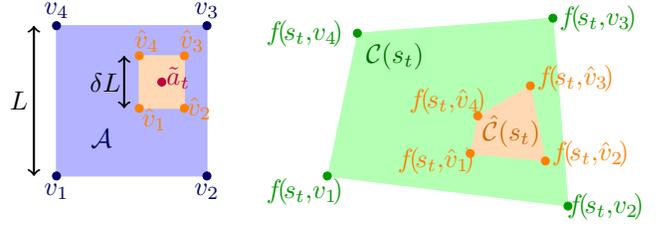
\begin{figure}[t]
\centering
\begin{tikzpicture}[scale = 1]
    \def\r{1.5pt}; 

    \filldraw[blue!30] (-3, -0.5) rectangle (-1, 1.5);
    \node at (-2.4, 0.) {\textcolor{blue!40!black}{$\mathcal{A}$}};
    \filldraw[blue!40!black] (-3, -0.5) circle (\r);
    \node at (-3, -0.7) {\textcolor{blue!40!black}{$v_1$}};
    \filldraw[blue!40!black] (-3, 1.5) circle (\r);
    \node at (-3, 1.7) {\textcolor{blue!40!black}{$v_4$}};
    \filldraw[blue!40!black] (-1, -0.5) circle (\r);
    \node at (-1, -0.7) {\textcolor{blue!40!black}{$v_2$}};
    \filldraw[blue!40!black] (-1, 1.5) circle (\r);
    \node at (-1, 1.7) {\textcolor{blue!40!black}{$v_3$}};
    \draw[thick, <->] (-3.3, -0.5) -- (-3.3, 1.5);
    \node at (-3.5, 0.5) {$L$};

    \filldraw[orange!30] (-1.9, 0.4) rectangle (-1.3, 1.1);
    \filldraw[orange] (-1.9, 0.4) circle (\r);
    \node at (-1.7, 0.25) {\textcolor{orange}{$\hat{v}_1$}};
    \filldraw[orange] (-1.9, 1.1) circle (\r);
    \node at (-1.8, 1.3) {\textcolor{orange}{$\hat{v}_4$}};
    \filldraw[orange] (-1.3, 0.4) circle (\r);
    \node at (-1.1, 0.4) {\textcolor{orange}{$\hat{v}_2$}};
    \filldraw[orange] (-1.3, 1.1) circle (\r);
    \node at (-1.2, 1.3) {\textcolor{orange}{$\hat{v}_3$}};
    \draw[thick, <->] (-2.1, 0.4) -- (-2.1, 1.1);
    \node at (-2.35, 0.75) {$\delta L$};

    \filldraw[purple] (-1.6, 0.75) circle (\r);
    \node at (-1.38, 0.8) {\textcolor{purple}{$\tilde{a}_t$}};

    \filldraw[green!30] (0.6, -0.5) -- (3.8, -0.9) -- (3.6, 1.6) -- (1., 1.4);
    \node at (1.5, 1.1) {\textcolor{green!40!black}{$\mathcal{C}(s_t)$}};
    
    \filldraw[green!60!black] (0.6, -0.5) circle (\r);
    \node at (0.3, -0.7) {\textcolor{green!60!black}{$f\!(\!s_t,\!v_1\!)$}};

    \filldraw[green!60!black] (3.8, -0.9) circle (\r);
    \node at (4.3, -0.9) {\textcolor{green!60!black}{$f\!(\!s_t,\!v_2\!)$}};

    \filldraw[green!60!black] (3.6, 1.6) circle (\r);
    \node at (4.2, 1.6) {\textcolor{green!60!black}{$f\!(\!s_t,\!v_3\!)$}};

    \filldraw[green!60!black] (1., 1.4) circle (\r);
    \node at (0.3, 1.4) {\textcolor{green!60!black}{$f\!(\!s_t,\!v_4\!)$}};

    \filldraw[orange!30] (2.5, -0.2) -- (3.5, -0.3) -- (3.3, 0.7) -- (2.6, 0.3);
    \node at (3.05, 0.1) {\textcolor{orange!80!black}{$\hat{\mathcal{C}}(s_t)$}};
   
    \filldraw[orange] (2.5, -0.2) circle (\r);
    \node at (2.05, -0.3) {\textcolor{orange}{$f\!(\!s_t,\!\hat{v}_1\!)$}};
    
    \filldraw[orange] (3.5, -0.3) circle (\r);
    \node at (4.1, -0.2) {\textcolor{orange}{$f\!(\!s_t,\!\hat{v}_2\!)$}};
    
    \filldraw[orange] (3.3, 0.7) circle (\r);
    \node at (3.9, 0.8) {\textcolor{orange}{$f\!(\!s_t,\!\hat{v}_3\!)$}};
    
    \filldraw[orange] (2.6, 0.3) circle (\r);
    \node at (2.2, 0.5) {\textcolor{orange}{$f\!(\!s_t,\!\hat{v}_4\!)$}};
    

\end{tikzpicture}
\caption{Illustration of the reduced search space \textcolor{orange}{$\hat{\mathcal{C}}(s_t)$} of \eqref{eq: shrunk reachable set approx} resulting from the reduced action space \textcolor{orange}{$\conv(\hat{v}_1, \hat{v}_2, \hat{v}_3, \hat{v}_4)$} of~\eqref{eq: shrunk action vertices} by a factor $\delta$ surrounding action prediction \textcolor{purple}{$\tilde{a}_t$}. Finding the best next state $s_{t+1}$ is much easier in the smaller set \textcolor{orange}{$\hat{\mathcal{C}}(s_t)$} than in the larger \textcolor{green!60!black}{$\mathcal{C}(s_t)$}.}
\label{fig: shrunk action set}
\end{figure}

To increase the precision of Algorithms~\ref{alg: proj} and \ref{alg: ref proj}, we propose to use a prediction of the best action to restrict the search space~\eqref{eq: reachable set approx} of projections~\eqref{eq: projection} and~\eqref{eq: ref proj} to a small region centered on this prediction instead of naively testing the entire action space.
To predict the action corresponding to each state transition, we leverage the capabilities of our diffusion models to predict both state and action sequences simultaneously. We use the predicted action $\tilde{a}_t$ as the center of an action search space shrunk by a factor $\delta \in (0, 1)$ and delimited by vertices
\begin{equation}\label{eq: shrunk action vertices}
    \hat{v}_i = \tilde{a}_t + \delta \Big(v_i - \frac{1}{m}\sum_{i\, =\, 1}^m v_i \Big)
\end{equation}
for all $i \in [\![1, m]\!]$. These vertices lead to a correspondingly reduced reachable set
\begin{equation}\label{eq: shrunk reachable set approx}
    \hat{\mathcal{C}}(s_t) = \conv\big( f(s_t, \hat{v}_1), ..., f(s_t, \hat{v}_m) \big)
\end{equation}
illustrated in Fig.~\ref{fig: shrunk action set}.

Projection $\mathcal{P}$ over polytope $\hat{\mathcal{C}}(s_t)$ of~\eqref{eq: shrunk reachable set approx} is a convex optimization
\begin{equation}\label{eq: action shrunk state projection}
    \mathcal{P}\big(\tilde{s}_{t+1}, \hat{\mathcal{C}}(s_t)\big) := \underset{\hat{c}\, \in\, \hat{\mathcal{C}}(s_t)}{\arg\min} \|\tilde{s}_{t+1} - \hat{c} \| = \sum_{i\, =\, 1}^m \lambda_i^* f(s_t, \hat{v}_i) 
\end{equation}
where
\begin{equation}\label{eq: convex optim}
    \lambda^* = \underset{\lambda_i\, \in\, [0,1]}{\arg\min} \Bigg\{ \Big\|\tilde{s}_{t+1} - \sum_{i\, =\, 1}^m \lambda_i f(s_t, \hat{v}_i)\Big\| \ \text{s.t.}\ \sum_{i\, =\, 1}^m \lambda_i = 1 \hspace{-0.5mm} \Bigg\}.
\end{equation}
Indeed, any point $\hat{c}$ in polytope $\hat{\mathcal{C}}(s_t)$ can be written as a convex combination of its vertices, i.e., $\hat{c} = \sum \lambda_i f(s_t, \hat{v}_i)$ with $\sum \lambda_i = 1$.
The optimal coefficients $\lambda^*$ of~\eqref{eq: convex optim} can then be used to approximate the corresponding action $a_t = \sum \lambda_i^* \hat{v}_i$. This approximation is exact for control affine dynamics as detailed in Remark~\ref{rmk: control affine} of Appendix~\ref{apx: inverse dynamics}.
We summarize this state-action projection in Algorithm~\ref{alg: SA proj}.

\begin{algorithm}[htb!]
\caption{State Action Trajectory Projection}\label{alg: SA proj}
\begin{algorithmic}[1]
\Require predicted trajectory $\tilde{\tau} = \big\{s_0, \tilde{s}_1,... \tilde{s}_H\big\} \in \mathcal{S}^{H+1}$, predicted actions $\tilde{\alpha} = \big\{ \tilde{a}_0, ..., \tilde{a}_{H-1} \big\} \in \mathcal{A}^H$.
\vspace{2mm}
\State $\tau = \big\{ s_0 \big\}$, $\alpha = \{\}$ \Comment{initialize the projections}
\For{$t=0$ to $H-1$}
\For{$i=1$ to $m$}
\State $\hat{v}_i = \tilde{a}_t + \delta \big(v_i - \frac{1}{m}\sum v_i \big)$ \Comment{extremal actions}
\EndFor
\State $\lambda^* = $ solve~\eqref{eq: convex optim}
\State $s_{t+1} = \sum \lambda^* f(s_t, \hat{v})$, $a_t = \sum \lambda^* \hat{v}$ \Comment{predictions}
\State $\tau \leftarrow \tau \cup \{s_{t+1}\}$, $\alpha \leftarrow \alpha \cup \{a_t\}$ \Comment{append to $\tau$, $\alpha$}
\EndFor
\State \textbf{return} $\tau$, $\alpha$
\end{algorithmic}
\end{algorithm}

\subsection{Using action predictions for admissible state prediction}

While Algorithm~\ref{alg: SA proj} uses the action predictions to guide the next state projection, we can use this action prediction to directly obtain an admissible next state. If we replace the predicted next state $\tilde{s}_{t+1}$ by the state actually obtained when applying the predicted action $\tilde{a}_t$ on the current state $s_t$ we obtain the following projector:
\begin{equation}\label{eq: A-proj}
    \mathcal{P}^\text{A}(s_t, \tilde{a}_t) := f(s_t, \tilde{a}_t).
\end{equation}
Then, replacing lines 3-6 of Algorithm~\ref{alg: SA proj} with $s_{t+1} = \mathcal{P}^\text{A}(s_t, \tilde{a}_t)$ and $a_t = \tilde{a}_t$ guarantees the admissibility of the projected trajectory $\tau$ while also removing convex optimization~\eqref{eq: convex optim}. 

\begin{figure}[t!]
    \centering
    \begin{tikzpicture}[scale=1, label distance=-0.8mm]
        
        \node[label = above :$s_t$] (0) at (0, 0){};
        \node at (0)[circle,fill,inner sep=2pt]{};

        \node[label = above :\textcolor{red!80}{$\tilde{s}_{t+1}$}] (1) at (5, 0){};
        \node at (1)[circle,fill,red!80,inner sep=2pt]{};
        \draw[very thick, decorate, decoration={brace,amplitude=5pt}] (4.6, 0.5) -- (5.4, 0.5) node[midway,yshift=-3em]{};

        \node[label = above :\textcolor{blue!70}{$\mathcal{P}^\text{A}$}] (2) at (6, -1){};
        \node at (2)[circle,fill,blue!70,inner sep=2pt]{};
        \draw[very thick, decorate,decoration={brace,amplitude=5pt}] (5.6, -0.5) -- (6.4, -0.5) node[midway,yshift=-3em]{};
        
        \draw[very thick, dashed, blue!70, ->] (0) to [out=-15, in=180] (2);
        \node at (3, -0.55) {\textcolor{blue!70}{$\tilde{a}_t$}};

        \node[label = below :\textcolor{green!70!black}{$\mathcal{P}^\text{SA}$}] (3) at (4.5, 0){};
        \node at (3)[circle,fill,green!70!black,inner sep=2pt]{};
        \draw[very thick, dashed, green!70!black, ->] (0) to [out=15, in=165] (3);
        \node at (2.25, 0.55) {\textcolor{green!70!black}{$\tilde{a}_t + \delta a_t$}};

        \def\xc{5};
        \def\yc{1.4};
        \def\r{0.28}; 
        \draw[black, thick] (\xc, \yc) circle (0.4);
        \node at (\xc +0.22, \yc) {$\boldsymbol{-}$};
        \node at (\xc, \yc-0.22) {$\boldsymbol{+}$};
        \draw[black, thick] (\xc-\r, \yc-\r) --  (\xc+\r, \yc+\r);
        \draw[black, thick] (\xc-\r, \yc+\r) --  (\xc+\r, \yc-\r);
        
        \node[draw=black, thick, rounded corners=4pt] (4) at (\xc-1.5, \yc) {$\pi_\theta$};
        \draw[->, very thick] (5, 0.7) -- (5, \yc-0.4);
        \draw[->, very thick] (6, -0.3) -- (6, \yc) -- (\xc+0.4, \yc);
        \draw[->, very thick] (\xc-0.4, \yc) to (4);
        \draw[->, very thick] (4) to (2.58, \yc) -- (2.58, 0.7);
    \end{tikzpicture}
    \vspace{-5mm}
    \caption{Illustration of projectors $\textcolor{blue!70}{\mathcal{P}^\text{A}} = f(s_t, \textcolor{blue!70}{\tilde{a}_t})$ and $\textcolor{green!70!black}{\mathcal{P}^\text{SA}} = f(s_t, \textcolor{green!70!black}{\tilde{a}_t + \delta a_t})$ with its feedback correction policy $\pi_\theta$ leveraging the open-loop error to generate a corrective action \textcolor{green!70!black}{$\delta a_t$} instead of discarding prediction \textcolor{red!80}{$\tilde{s}_{t+1}$} like $\textcolor{blue!70}{\mathcal{P}^\text{A}}$.}
    \label{fig: A-proj SA-proj}
\end{figure}
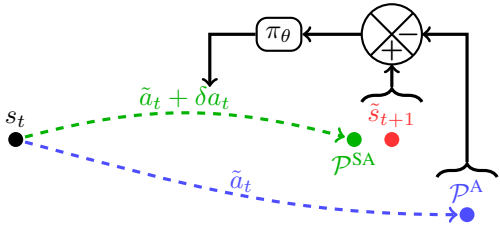

While projector~\eqref{eq: A-proj} entirely discards the next state prediction $\tilde{s}_{t+1}$, we can leverage its information to design a smarter projector. We propose to use a feedback correction term $\delta a_t$ on the action $\tilde{a}_t$ based on the distance between its corresponding next state $\mathcal{P}^A$ and the desired next state $\tilde{s}_{t+1}$, as illustrated in Fig.~\ref{fig: A-proj SA-proj}. Such a projector $\mathcal{P}^\text{SA}$ leverages the combined information of state and action predictions as
\begin{equation}\label{eq: SA-proj}
    \mathcal{P}^\text{SA}(s_t, \tilde{a}_t, \tilde{s}_{t+1}) := f(s_t, \tilde{a}_t + \delta a_t).
\end{equation}
To calculate the feedback correction $\delta a_t$ we prefer a small neural network $\pi_\theta$ to an optimization algorithm because we need $\mathcal{P}^{SA}$ to be differentiable and sufficiently fast to be incorporated into the training of our diffusion models. A model-based approach would not work either due to the black-box nature of $f$. Then, $\delta a_t := \pi_\theta\big(\tilde{s}_{t+1} - \mathcal{P}^\text{A}(s_t, \tilde{a}_t) \big)$. Given the dataset of trajectories $\mathcal{D}$, we can extract admissible triplets $(s_t, a_t, s_{t+1})$, sample a correction term $\delta a_t \sim \mathcal{N}(0, \sigma^2 I)$, correct next state $s^\delta_{t+1} := f(s_t, a_t + \delta a_t)$, and train $\pi_\theta$ as
\begin{equation*}
    \underset{\theta}{\min} \underset{\delta a_t \sim \mathcal{N}(0, \sigma^2 I)}{\mathbb{E}} \ \underset{(s_t, a_t, s_{t+1}) \sim \mathcal{D}}{\mathbb{E}} \big\| \pi_\theta\big(s^\delta_{t+1} - s_{t+1}\big) - \delta a_t \big\|^2.
\end{equation*}

\section{Incorporating Projections in Diffusion}\label{sec: DDAT}

In this section, we discuss how we can incorporate the different projectors of Section~\ref{sec: projections} into our diffusion model to learn diffusion policies generating dynamically admissible trajectories.

\subsection{Training with projections}

We introduce Algorithm~\ref{alg: DDAT} to train and sample admissible trajectories from our diffusion models.
Each training iteration begins with the sampling of a noise level $\sigma$ describing the magnitude of noising sequence $\varepsilon$. This noise sequence $\varepsilon$ is added to an admissible trajectory $\tau$ from dataset $\mathcal{D}$. The diffusion neural network predicts a denoised trajectory $\tilde{\tau}$ from the corrupted signal $\tau + \varepsilon$. If the noise level $\sigma$ is sufficiently low we project $\tilde{\tau}$ with one of the projectors $\mathcal{P}$ of Section~\ref{sec: projections}. The difference between this projection and the original noise-free trajectory $\tau$ is then used as a loss to update neural network $D_\theta$.
The curriculum deciding at which noise levels $\sigma$ projections can occur is discussed in Section~\ref{subsec: projection scheduling}.

After training $D_\theta$, we can use our model to plan a trajectory starting from a given initial state $s_0$. This inference phase starts with sampling a random sequence $\tau_0$ and imposing its first term to be $s_0$. Then, we iteratively process the trajectory through neural network $D_\theta$ and the projector $\mathcal{P}$ with which it was trained. After $N$ iterations of progressively reducing the noise level of our sample we obtain a noise-free admissible trajectory prediction starting from $s_0$. Since the projections are necessary during inference to generate admissible trajectories, we included the same projections into the training to match the inference.

\begin{algorithm}[htb!]
\caption{DDAT}\label{alg: DDAT}
\vspace{-2mm}
\begin{center}\textbf{Training}\end{center}
\vspace{-3mm}
\begin{algorithmic}[1]
\Require dataset of trajectories $\mathcal{D}$, projector $\mathcal{P}_\sigma$ of \eqref{eq: projection}, \eqref{eq: ref proj}, \eqref{eq: A-proj}, or \eqref{eq: SA-proj}.
\While{not converged}
\State $\sigma \sim \exp\big( \mathcal{N}(p_m, p_s^2) \big)$ \Comment{noise level, Appendix~\ref{apx: diffusion}.}
\State $\varepsilon \sim \mathcal{N}(0, \sigma^2 I)$ \Comment{noise sequence}
\State $\tau \sim \mathcal{U}(\mathcal{D})$ \Comment{training trajectory}
\State $\tilde{\tau} \leftarrow D_\theta(\tau + \varepsilon;\ \sigma)$ \Comment{denoised prediction}
\State $\tilde{\tau} \leftarrow \mathcal{P}_\sigma\big(\tilde{\tau} \big)$ \Comment{projection}
\State $\mathcal{L} = \| \tilde{\tau} - \tau \|^2$ \Comment{loss}
\State $\theta \leftarrow \theta - \nabla_\theta \mathcal{L}$ \Comment{gradient step}
\EndWhile
\end{algorithmic}
\vspace{-2mm}
\begin{center}\textbf{Inference}\end{center}
\vspace{-4mm}
\begin{algorithmic}[1]
\Require initial state $s_0$.
\State $\tau_0 \sim \mathcal{N}\big(0, \sigma_{0}^2 I\big)$ \Comment{initialize trajectory}
\State $\tau_0[0] \leftarrow s_0$ \Comment{known initial state}
\For{$i=0$ to $N-1$}
\State $\tau_{i+1} \leftarrow S_\theta(\tau_i; \sigma_i, \sigma_{i+1})$ \Comment{denoised prediction \eqref{eq: denoising operator}}
\State $\tau_{i+1}[0] \leftarrow s_0$ \Comment{known initial state}
\State $\tau_{i+1} \leftarrow \mathcal{P}_{\sigma_i}\big(\tau_{i+1} \big)$ \Comment{projection}
\EndFor
\State \textbf{return} $\tau_N$ \Comment{noise-free sample}
\end{algorithmic}
\end{algorithm}

While Algorithm~\ref{alg: DDAT} is written to predict trajectories of states, it can similarly generate states and actions in parallel and use projectors $\mathcal{P}_\sigma^\text{A}$ or $\mathcal{P}_\sigma^\text{SA}$ by substituting $\tau$ for $(\tau, \alpha)$, where $\alpha$ is the action sequence corresponding to $\tau$.

To use the reference projector $\mathcal{P}^\text{ref}$ of~\eqref{eq: ref proj} during training, we can obviously use the trajectories from the dataset as references to guide the projection. However, during inference, this dataset might be unavailable, or there might not be a trajectory matching closely the desired one. To solve this issue we leverage the imitation capabilities of diffusion models as illustrated in Fig.~\ref{fig: hopper proj} where the sampled trajectory matches exactly the desired trajectory. Hence, we use as reference the sampled trajectory before projection, i.e., line 6 of the Inference Algorithm~\ref{alg: DDAT} becomes $\tau_{i+1} \leftarrow \mathcal{P}^\text{ref}_{\sigma_i}\big(\tau_{i+1}, \tau^\text{ref} \leftarrow \tau_{i+1} \big)$.


\subsection{Projection curriculum}\label{subsec: projection scheduling}

Motivated by the different approaches employed in the literature, we now discuss how projections should be scheduled in our diffusion models. Indeed, work~\citep{Projected} uses projections after each iteration of inference, while work~\citep{CGD} only projects trajectories at the end of inference. In between these two extremes, works like~\citep{bastek2024physics, RyneBeeson, PhysDiff} realized that projections are only meaningful when the level of noise on the trajectory is sufficiently low.
We arrived to the same conclusion for scheduling projections during training where a meaningless projection of random states would create a large loss destabilizing the training of the diffusion neural network.

We thus created a projection curriculum to ensure projections only occur at sufficiently low noise levels $\sigma < \sigma_\text{min}$. We realized empirically that the sudden contribution of projections to the training loss as $\sigma$ passes below $\sigma_\text{min}$ can still destabilize training as trajectory projections can create a significant additional loss. We thus decided to create an intermediary transition region between the projection regime of $\sigma < \sigma_\text{min}$ and the noisy regime without projections of $\sigma > \sigma_\text{max}$. In this transition region the probability $p$ of projecting each state transition $(s_t, \tilde{s}_{t+1})$ grows linearly as $\sigma$ approaches $\sigma_\text{min}$.
More specifically, the curriculum for a projector $\mathcal{P}$ is a function of the noise level $\sigma$ as follows:
\begin{align}\label{eq: proj scheduling}
    &\mathcal{P}_\sigma(\tilde{s}_{t+1}, \cdot) := (1-b)\mathcal{P}(\tilde{s}_{t+1},\cdot) + b\tilde{s}_{t+1} \hspace{2mm} \text{with} \hspace{2mm} b\sim\mathcal{B}\big(p(\sigma) \big), \nonumber \\
    &\text{where} \hspace{2mm} p(\sigma) := \left\{ \begin{array}{ll}
    1 \quad &\text{if}\ \sigma > \sigma_\text{max},\\
    \displaystyle\frac{\sigma - \sigma_\text{min}}{\sigma_\text{max} - \sigma_\text{min}} \quad &\text{if}\ \sigma \in [\sigma_\text{min}, \sigma_\text{max}],\\
    0 \quad &\text{if}\ \sigma < \sigma_\text{min}, \end{array} \right.
\end{align}
where $b$ is a Bernoulli variable taking value $1$ with probability $p(\sigma)$ and $0$ otherwise. Thus, projection $\mathcal{P}$ occurs with probability $1 - p(\sigma)$.
In this transition regime, not projecting all the state transitions of a trajectory breaks the compounding effect and significantly reduces the total projection error as illustrated in Fig.~\ref{fig: DDAT scheme}.

\section{Implementation}\label{sec: implementation}

In this section, we provide the implementation details for training and evaluating our proposed diffusion models.

\subsection{Test environments}\label{subsec: datasets}

We deploy our approach on a set of different robotics environments:
\begin{itemize}
    \item Hopper (12 states, 3 actions),
    \item Walker (18 states, 6 actions),
    \item HalfCheetah (18 states, 6 actions),
    \item Quadcopter (17 states, 4 actions),
    \item Unitree GO1 and GO2 (37 states, 12 actions).
\end{itemize}
The Hopper, Walker, and HalfCheetah environments are from OpenAI Gym environments~\citep{Gym} with MuJoCo physics engine~\citep{mujoco}. Unitree GO1 and GO2 are quadruped robots also simulated with MuJoCo~\citep{mujoco_menagerie, mujoco_playground}. Finally, we also test DDAT in a quadcopter simulation~\citep{viljoen2024differentiable}. All these environments are \emph{underactuated} as a planned next state can be infeasible from the current state due to the robot's limited actuation. 
We collect hundreds of admissible trajectories from these simulation environments to create datasets  $\mathcal{D}$ as discussed in Appendix~\ref{apx: experiments}. 
For each of these environments we trained a variety of diffusion models predicting either only state trajectories, or only action sequences (conditioned on the given initial state), or predicting both states and actions. For each of these modalities we used Algorithm~\ref{alg: DDAT} to train models with various projection schemes best adapted to the specificities of each environment.

\subsection{Diffusion architecture}\label{subsec: implementation diffusion}

The objective of our diffusion models is to predict sequences, thus we chose a transformer backbone instead of the widely used U-Net~\citep{Diffuser, power2023sampling}. We implemented diffusion transformers (DiT)~\citep{DiT} adapted to trajectory predictions by~\citep{AdaptDiffuser} and relying on the notations and best practices of~\citep{karras2022elucidating}. More details on the diffusion process and our DiT architecture can be found in Appendix~\ref{apx: diffusion}.

To incorporate our trajectory projections into the training of our diffusion policy $D_\theta$ as illustrated in Fig.~\ref{fig: DDAT scheme}, we need our projectors to be differentiable, which is accomplished by using \textit{cvxpylayers}~\citep{cvxpylayers2019} to solve the convex optimization~\eqref{eq: convex optim}. Indeed, neural network training requires to propagate gradients of the loss $\mathcal{L} = \|\mathcal{P}_\sigma(D_\theta(\tau + \varepsilon)) - \tau\|$ to $D_\theta$ and thus have to go through projector $\mathcal{P}_\sigma$. Since the black-box simulator $f$ is typically not differentiable and is a part of the projections, the gradients are not propagated perfectly through $\mathcal{P}$.

The inference of Algorithm~\ref{alg: DDAT} is performed on a batch of trajectories for each initial state $s_0$ and is followed by a selection phase to pick the best trajectory according to a desired metric. For most of our robots we select the predicted trajectory satisfying known state constraints for the longest time interval. For instance, on the Hopper, Walker, and Unitree this is the time before falling.

\subsection{Evaluation metrics}\label{subsec: inverse dynamics}

We will evaluate our approach by answering the following questions.
\begin{enumerate}[label=\textbf{Q\arabic*}]
    \item\label{Q: DAD better} How close to dynamical admissibility are the different DDAT models?
    \item\label{Q: projection inference timeline} Can a diffusion model generate higher quality trajectories with projections starting at the beginning of inference, after inference, or progressively during inference?
    \item\label{Q: projection in training} Does training diffusion models with a projector help them generate better trajectories than only using the projector during inference?
    \item\label{Q: diffusion on S, SA, A} Is it easier to generate high-quality admissible trajectories when the diffusion model predicts the states, the actions, or both?
\end{enumerate}

To quantify the dynamical admissibility of trajectories we need to measure the distance between each predicted next state $\tilde{s}_{t+1}$ and the closest admissible state belonging to the reachable set of $s_t$. We thus need a ground-truth inverse dynamics model, used \emph{only for evaluation} and defined as
\begin{equation}\label{eq: ID}
    I\!D(s_t, s_{t+1}) := \underset{a\, \in\, \mathcal{A}}{\arg\min} \big\{ \big\| s_{t+1} - f(s_t, a) \big\| \big\},
\end{equation}
where the minimum is well-defined if $\mathcal{A}$ is compact and $f$ is continuous. When $s_{t+1}$ is reachable from $s_t$, the minimum of the norm in \eqref{eq: ID} is $0$ and $f\big( s_t, I\!D(s_t, s_{t+1}) \big) = s_{t+1}$.
Given a sequence of states $\tilde{\tau} = \big\{s_0, \tilde{s}_1,... \tilde{s}_H\big\} \in \mathcal{S}^{H+1}$ we autoregressively apply the inverse dynamics model to find the closest admissible next state to the prediction as detailed in Algorithm~\ref{alg: ID traj} located in Appendix~\ref{apx: inverse dynamics}.
Solving optimization problem~\eqref{eq: ID} with sufficient precision to make $I\!D$ a ground-truth model is challenging due to the black-box nature of $f$. We discuss different approaches to solve~\eqref{eq: ID} in Appendix~\ref{apx: inverse dynamics}.
Equipped with this inverse dynamics model, we use the following two criteria to quantify the admissibility of trajectories.

\begin{definition}\label{def: SAE}
    The \emph{statewise admissibility error (SAE)} is the distance between next state prediction and closest admissible next state obtained from the inverse dynamics:
    \begin{equation}\label{eq: SAE}
        SAE(s_t, \tilde{s}_{t+1}) := \big\| \tilde{s}_{t+1} - f\big(s_t, I\!D(s_t, \tilde{s}_{t+1})\big) \big\|.
    \end{equation}
\end{definition}

\begin{definition}
    The \emph{cumulative admissibility error (CAE)} is the distance between a predicted trajectory $\tilde{\tau} = \big\{\tilde{s}_0, \tilde{s}_1, ..., \tilde{s}_H\big\}$ and its closest admissible projection
    \begin{equation}\label{eq: CAE}
        CAE(\tilde{\tau}) := \big\| \big\{\tilde{s}_1, ..., \tilde{s}_H\big\} - \big\{s_1, ..., s_H\big\} \big\|,
    \end{equation}
    where $s_0 := \tilde{s}_0$ and $s_{t+1} = f\big(s_t, I\!D(s_t, \tilde{s}_{t+1}) \big)$.
\end{definition}

The CAE of a trajectory will be larger than its average SAE due to compounding errors in the CAE.

\textit{Model nomenclature:} To distinguish the models evaluated, their name starts with a letter $S$, $A$, or $SA$ depending on whether they predict states, actions or both. The second part denotes the projector used, i.e., $\mathcal{P}$~\eqref{eq: projection}, $\mathcal{P}^\text{ref}$~\eqref{eq: ref proj}, $\mathcal{P}^\text{A}$~\eqref{eq: A-proj}, or $\mathcal{P}^\text{SA}$~\eqref{eq: SA-proj}. The projector can take two subscripts: \textsubscript{$\sigma$} if a projection curriculum is used and \textsubscript{$I$} if the projector is only used at inference, it is otherwise assumed to be also used during training. Finally, conditioning of models on a variable $v$ is denoted by $(\cdot | v)$.

\section{Experiments}\label{sec: simulations}

We will now answer \ref{Q: DAD better} in Section~\ref{subsec: Q1}, \ref{Q: projection inference timeline} in Section~\ref{subsec: Q2}, \ref{Q: projection in training} in Section~\ref{subsec: Q3} and \ref{Q: diffusion on S, SA, A} in Section~\ref{subsec: Q4}.

\subsection{Generating admissible trajectories}\label{subsec: Q1}

To answer \ref{Q: DAD better} we generate state trajectories with various models and evaluate the distance to admissibility of these trajectories by comparing their SAE and CAE.
Since the inverse dynamics~\eqref{eq: ID} are used as verification tool, none of the state models can achieve a smaller error than the precision of the inverse dynamics. This precision is measured by evaluating $I\!D$ on a dataset of admissible trajectories. On the other hand, trajectories generated with projectors $\mathcal{P}^\text{A}$ of~\eqref{eq: A-proj} and $\mathcal{P}^\text{SA}$ of~\eqref{eq: SA-proj} are automatically admissible since generated by their corresponding action sequences.

\begin{table}[htb!]
    \definecolor{tablegray}{RGB}{90, 98, 115}
    \centering
    \addtolength{\tabcolsep}{-1pt}
    \begin{tabular}{cccc}
        \toprule
         System & Model & SAE ($\downarrow$) &  CAE ($\downarrow$) \\ \midrule
         \multirow{6}{*}{Hopper} & $S$ & $1.1 \text{e}{-1}$ & $1.7$ \\
          & $S\mathcal{P}_{I}$ & $8.5 \text{e}{-3}$ & $4.8 \text{e}{-1}$ \\
          & $SA\mathcal{P}_{I}$ & $4.2 \text{e}{-4}$ & $6.5 \text{e}{-2}$ \\
          & $S\mathcal{P}_\sigma^\text{ref}$ & $7.5 \text{e}{-3}$ & $6.4 \text{e}{-1}$ \\
          & $SA\mathcal{P}_\sigma^\text{A, SA}$ & $\mathbf{0}$ & $\mathbf{0}$ \\
          & \textcolor{tablegray}{$I\!D$} & \textcolor{tablegray}{$6.9 \text{e}{-5}$} & \textcolor{tablegray}{$9.5 \text{e}{-4}$} \\[2mm]
         \hdashline \\[-2mm]
         \multirow{5}{*}{Walker} & $SA$ & $3.1 \text{e}{-1}$ & $3.9$ \\
          & $SA\mathcal{P}_{I}$ & $5.4 \text{e}{-4}$ & $3.5 \text{e}{-1}$ \\
          & $SA\mathcal{P}_\sigma^\text{ref}$ & $7.1 \text{e}{-4}$ & $3.5 \text{e}{-1}$ \\
          & $SA\mathcal{P}_\sigma^\text{A, SA}$ & $\mathbf{0}$ & $\mathbf{0}$ \\
          & \textcolor{tablegray}{$I\!D$} & \textcolor{tablegray}{$1.3 \text{e}{-5}$} & \textcolor{tablegray}{$6.2 \text{e}{-2}$} \\[2mm] \hdashline \\[-2mm]
         \multirow{5}{*}{HalfCheetah} & $SA$ & $2.8 \text{e}{-1}$ & $4.6$ \\
          & $SA\mathcal{P}_{I}$ & $4.8 \text{e}{-3}$ & $1.99$ \\
          & $SA\mathcal{P}_\sigma^\text{ref}$ & $4.4 \text{e}{-4}$ & $2.1$ \\
          & $SA\mathcal{P}_\sigma^\text{A, SA}$ & $\mathbf{0}$ & $\mathbf{0}$ \\
          & \textcolor{tablegray}{$I\!D$} & \textcolor{tablegray}{$5.5 \text{e}{-12}$} & \textcolor{tablegray}{$3.1 \text{e}{-2}$} \\[2mm]
         \hdashline \\[-2mm]
         \multirow{6}{*}{Quadcopter} & $S$  & $2.55 \text{e}{-1}$ & $9.57 \text{e}{-1}$ \\
          & $S\mathcal{P}_\sigma^\text{ref}$ & $4.77 \text{e}{-4}$ & $4.88 \text{e}{-4}$ \\
          & $SA$ & $3.24 \text{e}{-1}$ & $1.48$  \\
          & $SA\mathcal{P}_\sigma^\text{ref}$ & $1.22 \text{e}{-6}$ & $1.68 \text{e}{-5}$ \\
          & $SA\mathcal{P}_\sigma^{A, SA}$ & $\mathbf{0} $ & $\mathbf{0}$ \\
          & \textcolor{tablegray}{$I\!D$} & \textcolor{tablegray}{$1.14 \text{e}{-17}$} & \textcolor{tablegray}{$1.04 \text{e}{-16}$} \\ \bottomrule
    \end{tabular}
    \caption{Distance to admissibility of $100$ trajectories generated by diffusion models. The errors are calculated with the inverse dynamics models ($I\!D$) whose precision limit that of the models. Applying any projection scheme decreases the admissibility errors by orders of magnitude on all robots.}
    \label{tab: Q1}
\end{table}

Table~\ref{tab: Q1} shows that applying any of our projection schemes to the diffusion models reduces both their statewise and cumulative admissibility error, even when the projection is only applied at inference. Models predicting both states and actions tend to have smaller admissibility errors than models predicting only states. On the Hopper and Walker our state-action models are only one order of magnitude away from the inverse dynamics precision whereas the HalfCheetah and Quadcopter required extremely precise inverse dynamics model. Thus Table~\ref{tab: Q1} allows to answer \ref{Q: DAD better}.

\subsection{Projection curriculum during inference}\label{subsec: Q2}

We will now address \ref{Q: projection inference timeline} motivated by the discussion of Section~\ref{subsec: projection scheduling} on the best projection curriculum to adopt. 
Work~\citep{Projected} uses projections at each step of the inference process, which we implement with curriculum~\eqref{eq: proj scheduling} parametrized by $\sigma_{min} = \sigma_{max} = 80$, highest noise level of inference to ensure a projection takes place at each iteration.
To reproduce the setting of~\citep{CGD} with post-inference projection we set $\sigma_{min} = \sigma_{max} = 0.0021$. To emulate the approaches of \citep{bastek2024physics, RyneBeeson, PhysDiff} we select $\sigma_{min} = 0.0021$ and $\sigma_{max} = 0.2$.

\begin{figure*}[t!]
    \begin{subfigure}[]{0.32\linewidth}
        \includegraphics[width=\linewidth]{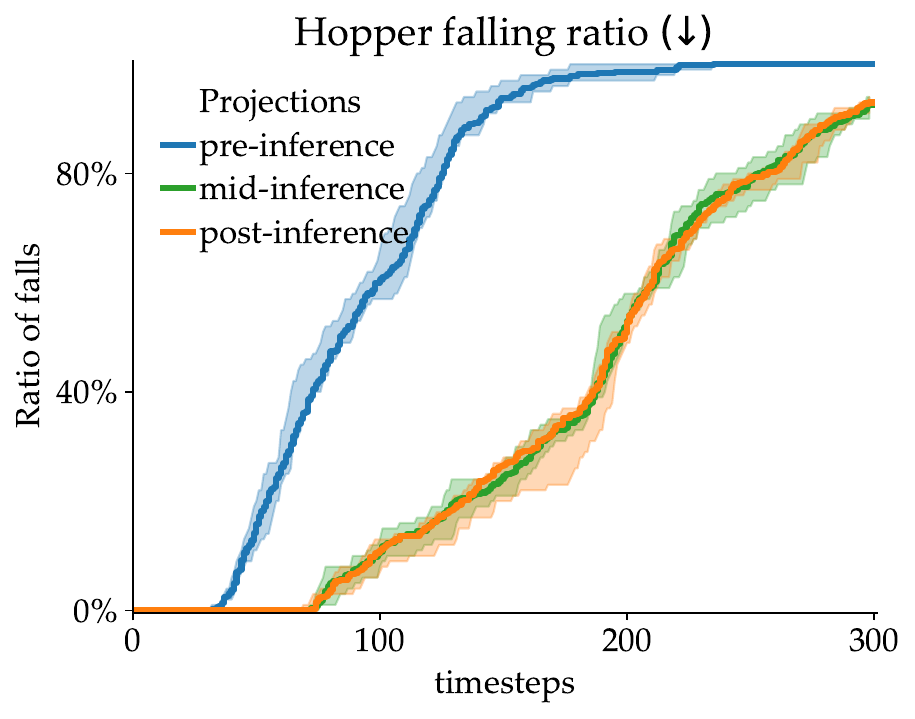}
        \caption{Hopper $SA\mathcal{P}_\sigma$ with three projection curricula. Starting projections at the beginning of inference (\textcolor{blue!80}{blue}) makes the Hopper fall much earlier, nearly $90\%$ at 150 timesteps compared to $20\%$ for the other curricula, which perform equivalently.}
        \label{fig: Q2 Hopper SA fall}
    \end{subfigure}\hfill
    \begin{subfigure}[]{0.32\linewidth}
        \includegraphics[width=\linewidth]{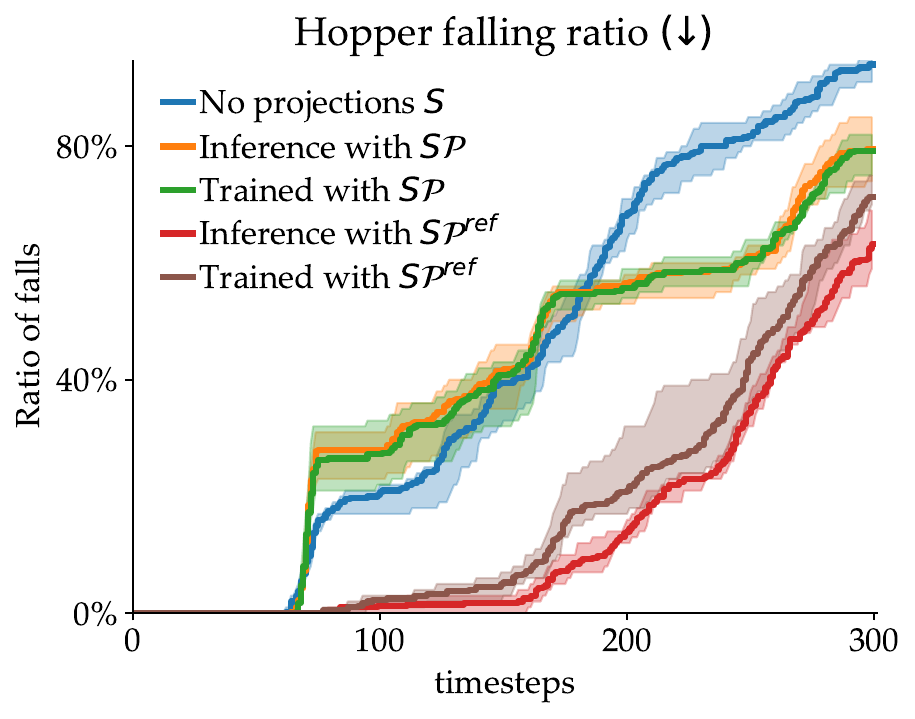}
        \caption{Fall ratios for the Hopper state models \textcolor{blue!80}{$S$}, \textcolor{orange!90!black}{$S\mathcal{P}_I$}, \textcolor{green!60!black}{$S\mathcal{P}$}, \textcolor{red!80!black}{$S\mathcal{P}_{I\sigma}^\text{ref}$} and \textcolor{brown!60!black}{$S\mathcal{P}_\sigma^\text{ref}$}. The falling ratio is consistently higher without projections and with the naive projector $\mathcal{P}$ of Algorithm~\ref{alg: proj} compared to the reference projector $\mathcal{P}^\text{ref}$ of Algorithm~\ref{alg: ref proj}. Training with projections or only using projections at inference has no significant impact.}
        \label{fig: Hopper falling ratio}
    \end{subfigure}\hfill
    \begin{subfigure}[]{0.32\linewidth}
        \centering
        \includegraphics[width=\linewidth]{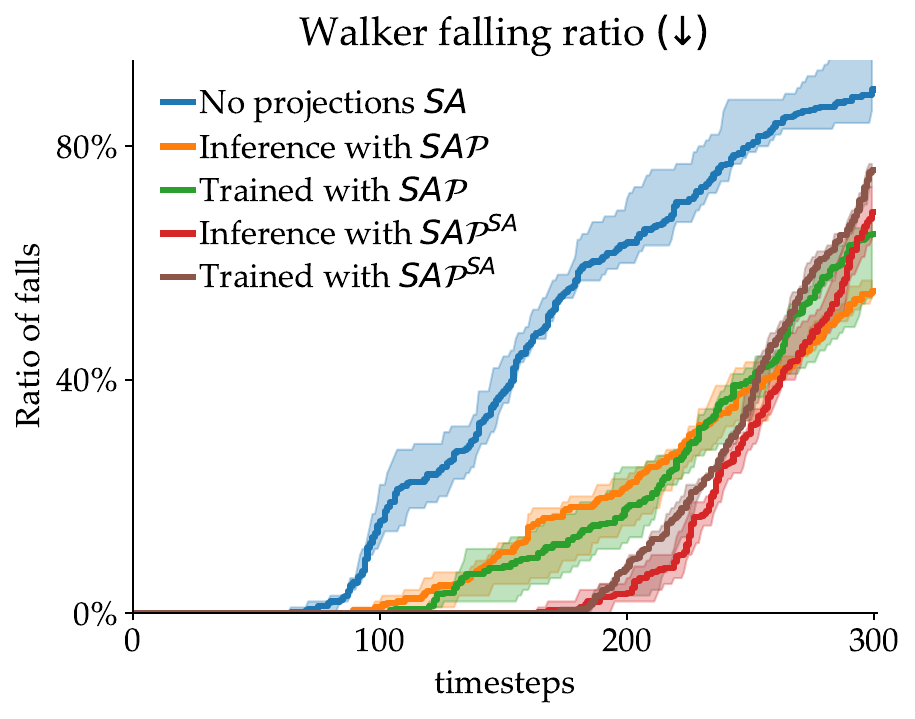}
        \caption{Fall ratios for the Walker state-action models \textcolor{blue!80}{$SA$}, \textcolor{orange!90!black}{$SA\mathcal{P}_I$}, \textcolor{green!60!black}{$SA\mathcal{P}$}, \textcolor{red!80!black}{$SA\mathcal{P}_{I\sigma}^\text{SA}$} and \textcolor{brown!60!black}{$SA\mathcal{P}_\sigma^\text{SA}$}. Without projections, the Walker falls much earlier than using any projections. The state-action projector $\mathcal{P}^\text{SA}$ of~\eqref{eq: SA-proj} prevents the Walker from falling for longer than projection $\mathcal{P}$ of \eqref{eq: action shrunk state projection}. Training with projections has no significant impact.}
        \label{fig: Walker falling ratio}
    \end{subfigure}\vspace{4mm}
    \caption{Ratios of trajectories deployed open-loop having fallen at a given timestep for the Hopper and Walker. The shade is the maximum and minimum number of trajectories having fallen at each timestep over 5 runs of 100 trajectories each.}
   \label{fig: falling ratios}
   \vspace{2mm}
\end{figure*}

\begin{table*}[htb!]
\centering
\begin{tabular}{ccccccccc}
    \toprule
     \multicolumn{3}{c}{parameters} & \multicolumn{4}{c}{Hopper $SA\mathcal{P}_\sigma$} & \multicolumn{2}{c}{Quadcopter $SA\mathcal{P}_\sigma^\text{SA}$} \\
     \cmidrule(lr){1-3} \cmidrule(lr){4-7} \cmidrule(lr){8-9}
     Model & $\sigma_{min}$ & $\sigma_{max}$ &  SAE ($\downarrow$) &  CAE ($\downarrow$) & survival\% ($\uparrow$) & reward ($\uparrow$) & survival\% ($\uparrow$) & reward ($\uparrow$) \\ 
     \cmidrule(lr){1-3} \cmidrule(lr){4-7} \cmidrule(lr){8-9}
     pre  & $80$ & $80$ & $4.92 \text{e}{-4}$ & $5.57 \text{e}{-2}$ & $27 \pm 10$ & $131 \pm 46$ & $85 \pm 20$ & $16 \pm 14$ \\
     mid  & $0.0021$ & $0.2$ & $\mathbf{1.41 \text{e}{-4}}$ & $6.13 \text{e}{-2}$ & $\mathbf{65 \pm 21}$ & $\mathbf{364 \pm 130}$ & $\mathbf{90 \pm 21}$ & $\mathbf{93 \pm 34}$ \\
     post & $0.0021$ & $0.0021$ & $1.45\text{e}{-4}$ & $\mathbf{4.82 \text{e}{-2}}$ & $63 \pm 24$ & $345 \pm 152$ & $\mathbf{90 \pm 21}$ & $\mathbf{93 \pm 34}$ \\ \bottomrule
\end{tabular}
\caption{Distance to admissibility of generated trajectories projected either at each step of inference (pre), gradually during inference (mid), or only after inference (post). The results are averaged over 500 trajectories and clearly demonstrate that projections at high noise level (pre) hurt significantly the quality of the trajectories in terms of survival and reward. The admissibility is only marginally affected by the projection curriculum. The difference between post-inference projections and mid-inference projections is not significant. 
The Quadcopter trajectories are projected with $SA\mathcal{P}_\sigma^\text{SA}$ and hence are all admissible.}
\label{tab: admissibility Q2}
\end{table*}

\begin{figure}[t!]
    \centering
    \includegraphics[width=\linewidth]{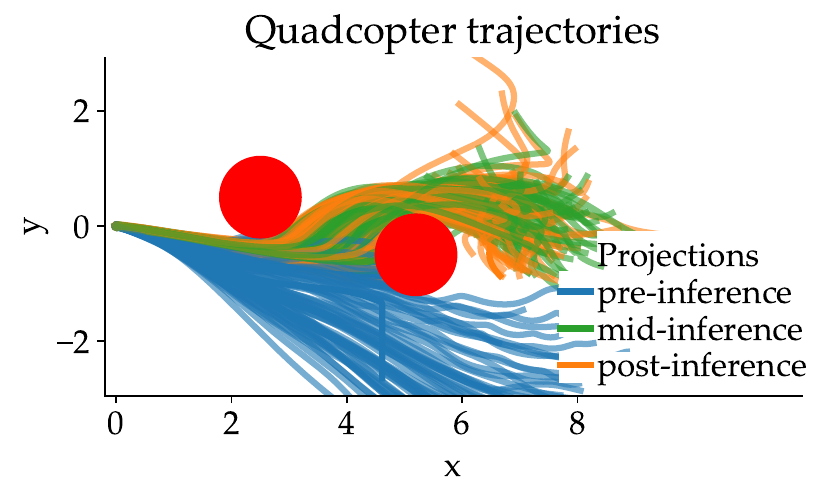}
    \hspace{-4mm}
    \caption{Quadcopter trajectories generated by the $SA\mathcal{P}_\sigma^\text{SA}$ model with projections starting either \textbf{\textcolor{blue!70}{pre-inference}}, or \textbf{\textcolor{green!50!black!80}{mid-inference}} or \textbf{\textcolor{orange!95!black!80}{post-inference}}. The desired behavior exhibited in Fig.~\ref{fig: quad slalom} slaloms between the \textbf{\textcolor{red}{obstacles}}. None of the trajectories generated by the pre-inference model pass the obstacles, whereas \textbf{\textcolor{green!50!black!80}{$81\%$}} and \textbf{\textcolor{orange!95!black!80}{$79\%$}} of the trajectories generated by the other two projection curricula pass the slalom. Hence, projections at the beginning of inference strongly reduce the quality of the samples.}
    \label{fig: quad trajs curriculum}
\end{figure}

Table~\ref{tab: admissibility Q2} and Figures~\ref{fig: Q2 Hopper SA SAE} and \ref{fig: Q2 Hopper SA CAE} show no significant difference in terms of admissibility based on the projection curriculum. However, the quality of the trajectories is significantly reduced by projections at high noise level as shown in Table~\ref{tab: admissibility Q2} and Fig.~\ref{fig: Q2 Hopper SA fall} and \ref{fig: quad trajs curriculum}. This answers \ref{Q: projection inference timeline}.

\subsection{Training with projections}\label{subsec: Q3}

\begin{figure*}[htb!]
    \captionsetup[subfigure]{justification=Centering}
    \begin{subfigure}[]{0.32\linewidth}
        \includegraphics[width=\linewidth]{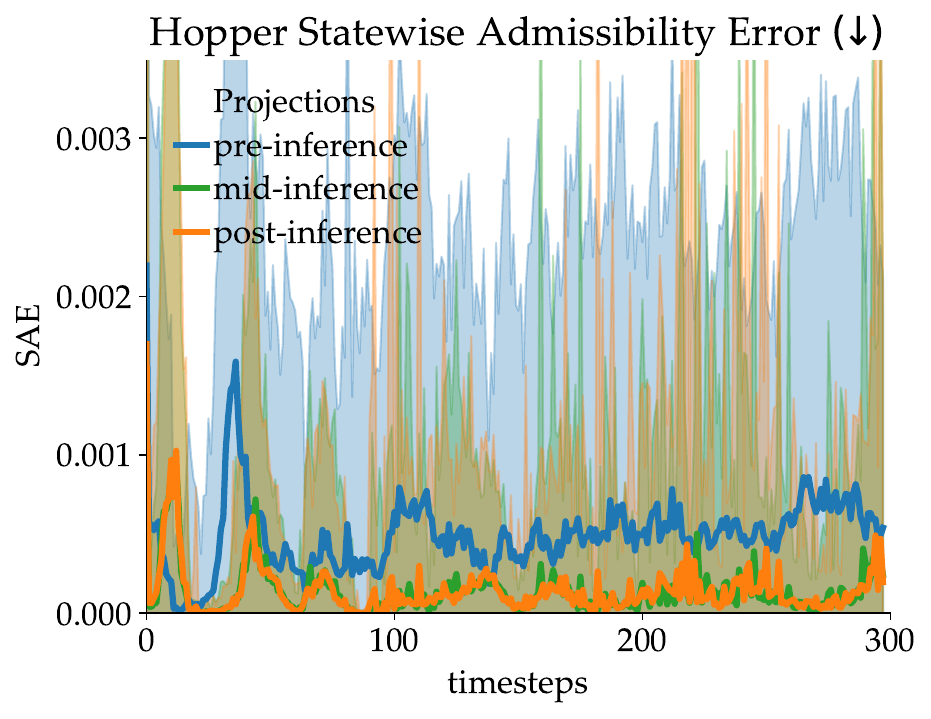}
        \caption{Hopper SAE for state-action models with different projections curricula.}
        \label{fig: Q2 Hopper SA SAE}
    \end{subfigure}\hfill
    \begin{subfigure}[]{0.32\linewidth}
        \includegraphics[width=\linewidth]{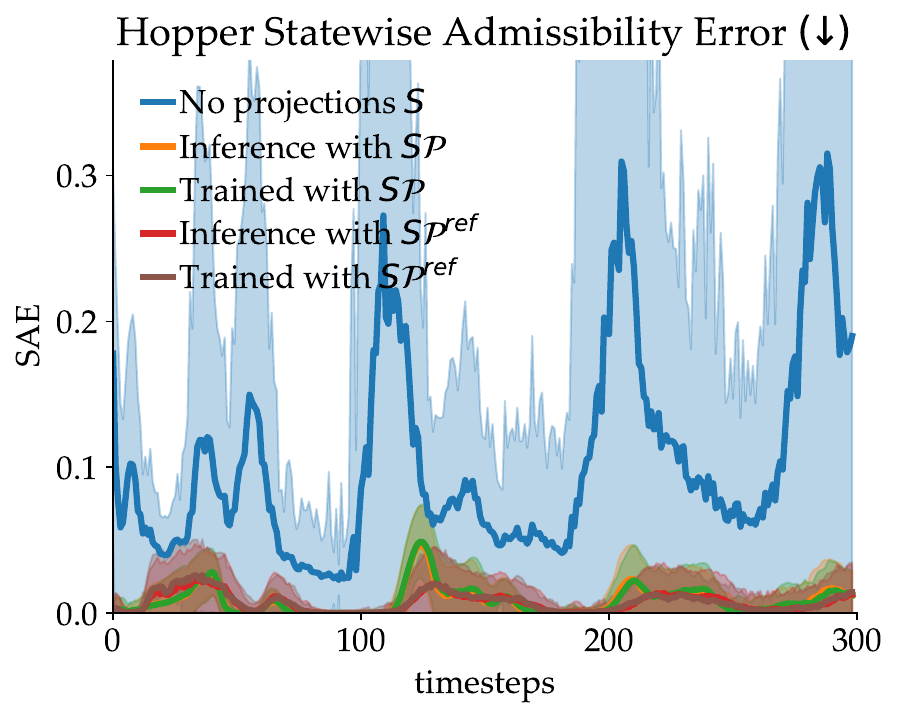}
        \caption{Hopper SAE for state models.}
        \label{fig: Q1 Hopper S SAE}
    \end{subfigure}\hfill
    \begin{subfigure}[]{0.32\linewidth}
        \includegraphics[width=\linewidth]{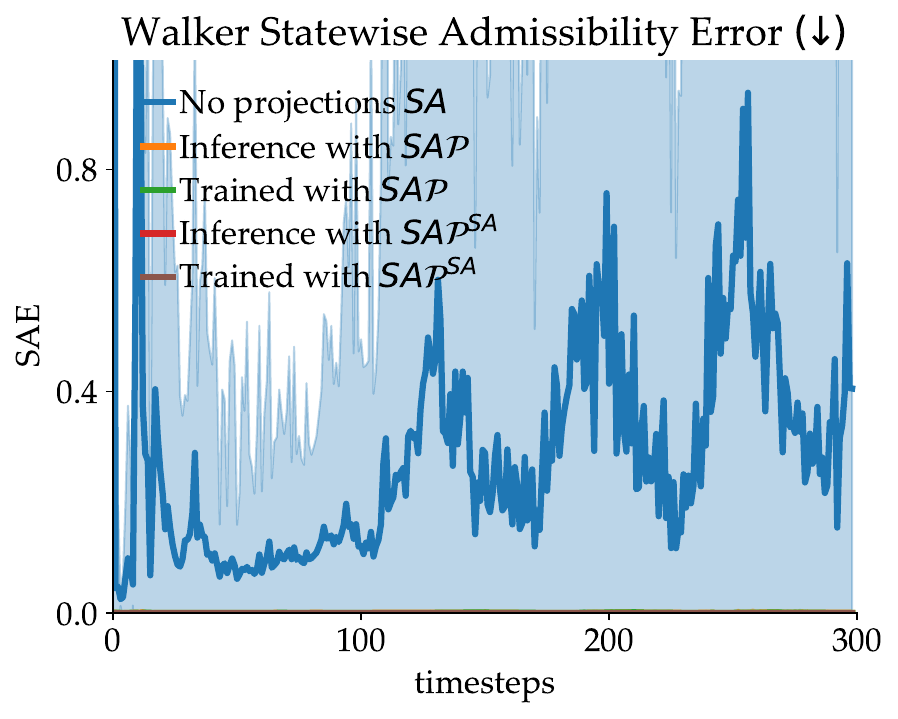}
        \caption{Walker SAE for state-action models.}
        \label{fig: Q1 Walker SA SAE}
    \end{subfigure}\\[5mm]

    \begin{subfigure}[]{0.32\linewidth}
        \includegraphics[width=\linewidth]{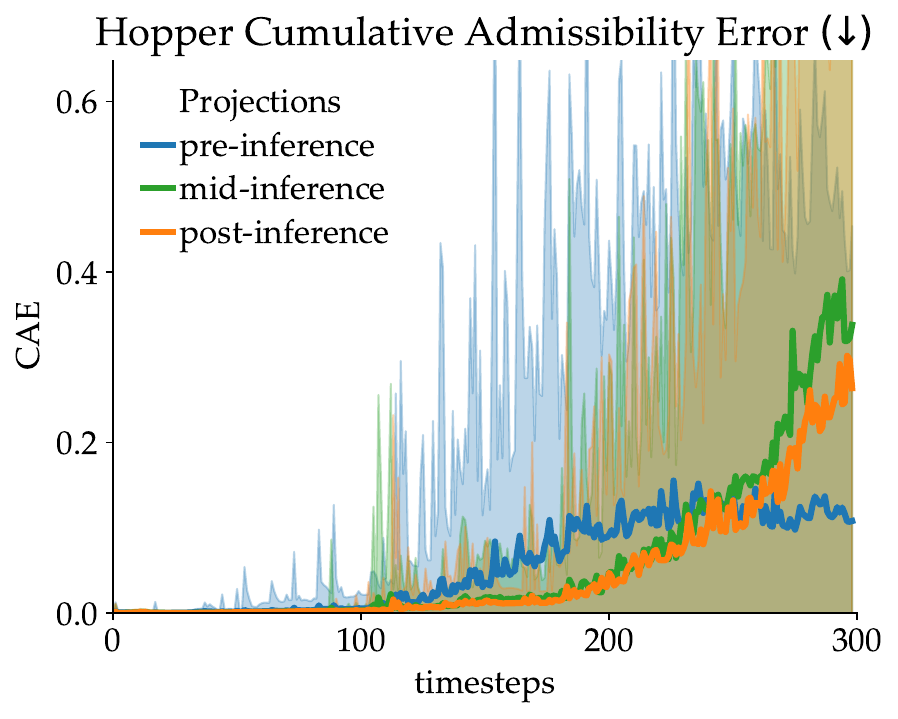}
        \caption{Hopper CAE for state-action models with different projections curricula.}
        \label{fig: Q2 Hopper SA CAE}
    \end{subfigure}\hfill
    \begin{subfigure}[]{0.32\linewidth}
        \includegraphics[width=\linewidth]{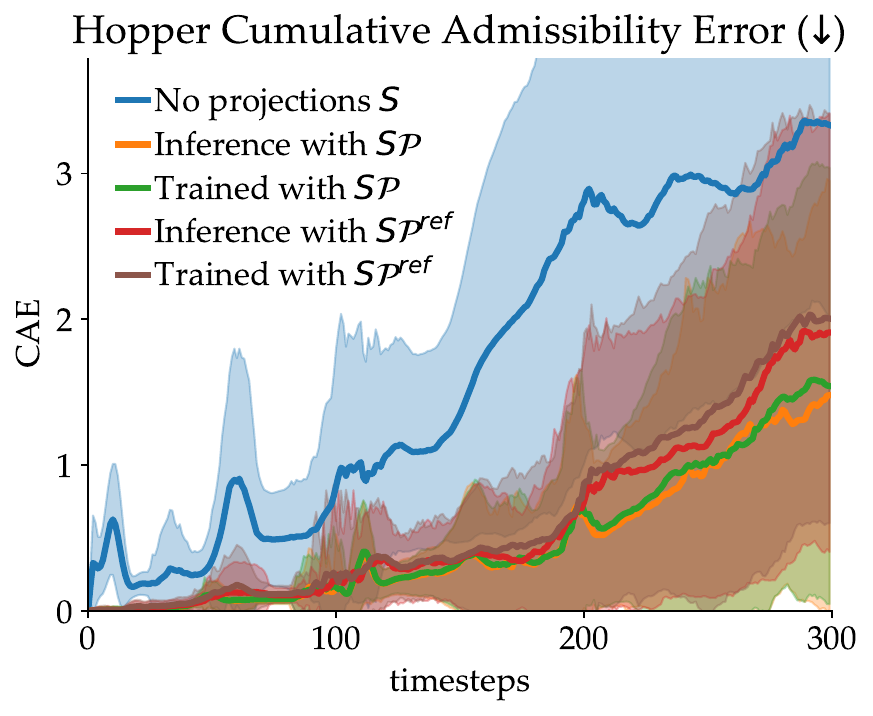}
        \caption{Hopper CAE for state models.}
        \label{fig: Q1 Hopper S CAE}
    \end{subfigure}\hfill
    \begin{subfigure}[]{0.32\linewidth}
        \includegraphics[width=\linewidth]{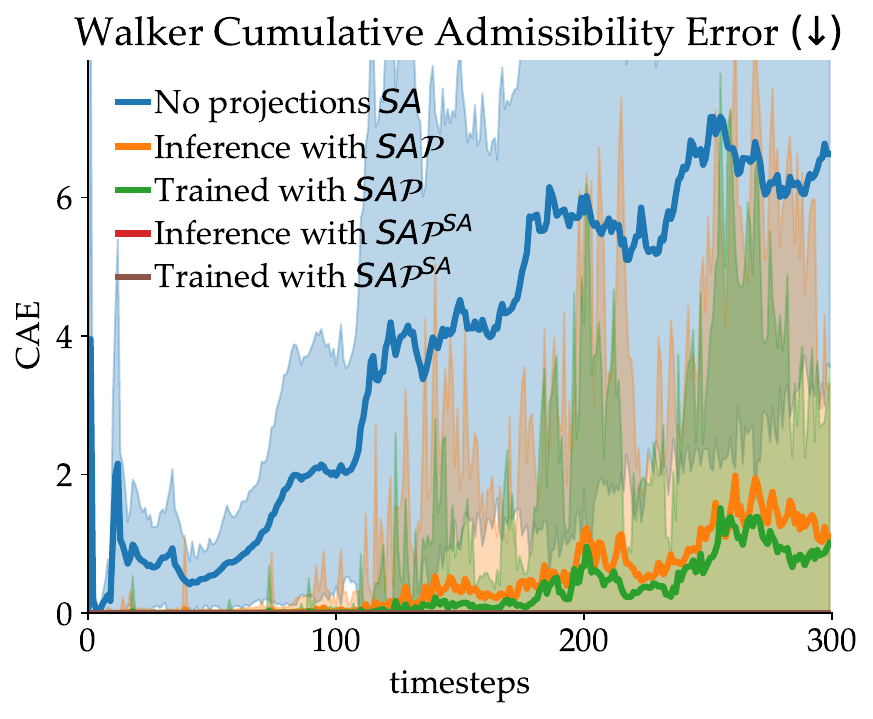}
        \caption{Walker CAE for state-action models.}
        \label{fig: Q1 Walker SA CAE}
    \end{subfigure}
    \vspace{5mm}
    \caption{Statewise (SAE) and Cumulative (CAE) Admissibility Errors for different Hopper and Walker models collected over 5 runs of 100 initial conditions with 8 samples each. The shade shows the standard deviation. Figures (\subref{fig: Q1 Hopper S SAE}) (\subref{fig: Q1 Walker SA SAE}), (\subref{fig: Q1 Hopper S CAE}) and (\subref{fig: Q1 Walker SA CAE}) show that projections drastically improve the admissibility of models. Figures (\subref{fig: Q2 Hopper SA SAE}) and (\subref{fig: Q2 Hopper SA CAE}) show small difference in admissibility when projecting during inference}
    \label{fig:Hopper CAE SAE}
    \vspace{2mm}
\end{figure*}

Let us now answer \ref{Q: projection in training} on whether incorporating projections during training helps diffusion models generate better samples.
Our intuition was that projections during training would better prepare the diffusion models for the projections at inference. However, Figures~\ref{fig: Hopper falling ratio} and \ref{fig: Walker falling ratio} both show little difference in trajectory quality between the models trained with projections and those using projections only at inference. The only significant quality difference is caused by employing different projectors.
In terms of admissibility, Figures~\ref{fig: Q1 Hopper S SAE}, \ref{fig: Q1 Walker SA SAE}, \ref{fig: Q1 Hopper S CAE}, and \ref{fig: Q1 Walker SA CAE} show no significant difference between models trained with projections and those only using them at inference.
Despite our best effort to incorporate projections into the training of diffusion models, projections still disrupt training by significantly increasing the training loss. Thus our answer to \ref{Q: projection in training} is negative, training with projections is not helpful.

\begin{table*}[htb!]
\definecolor{tablegray}{RGB}{90, 98, 115}
\centering
\begin{tabular}{ccccccccc}
    \toprule
    \multirow{2}{*}{Model} & \multicolumn{2}{c}{Hopper} & \multicolumn{2}{c}{Walker} & \multicolumn{2}{c}{HalfCheetah} & \multicolumn{2}{c}{Quadcopter} \\
     \cmidrule(lr){2-3} \cmidrule(lr){4-5} \cmidrule(lr){6-7} \cmidrule(lr){8-9}
     & survival ($\uparrow$) &  reward ($\uparrow$) & survival ($\uparrow$) &  reward ($\uparrow$) & survival ($\uparrow$) &  reward ($\uparrow$) & survival ($\uparrow$) &  reward ($\uparrow$) \\
     \cmidrule(lr){1-1} \cmidrule(lr){2-3} \cmidrule(lr){4-5} \cmidrule(lr){6-7} \cmidrule(lr){8-9}
     MPPI~\cite{MPPI}  & $10 \pm 3$ & $85 \pm 59$ & $0 \pm 0$ & $1 \pm 6$ & $0 \pm 0$ & $-33 \pm 37$ & $0 \pm 0$ & $-639 \pm 225$\\
     \textcolor{tablegray}{data} & \textcolor{tablegray}{$100 \pm 0$} & \textcolor{tablegray}{$563 \pm 21$} & \textcolor{tablegray}{$100 \pm 0$} & \textcolor{tablegray}{$995 \pm 80$} & \textcolor{tablegray}{$100 \pm 0$} & \textcolor{tablegray}{$611 \pm 30$} & \textcolor{tablegray}{$100 \pm 0$}  & \textcolor{tablegray}{$117 \pm 5.7$} \\
     $SA$ & $50 \pm 25$ & $294 \pm 190$ & $58 \pm 22$ & $333 \pm 184$ & $87 \pm 15$ & $226 \pm 88$ & $65 \pm 31$ & $32 \pm 28$ \\
     $SA\mathcal{P}_{I}$ & $67 \pm 22$ & $369 \pm 141$ & $84 \pm 18$ & $516 \pm 177$ & $94 \pm 10$ & $318 \pm 104$ & $66 \pm 29$ & $43 \pm 28$ \\ 
     $SA\mathcal{P}_\sigma^\text{ref}$ & $\mathbf{72 \pm 19}$ & $\mathbf{410 \pm 153}$ & $83 \pm 11$ & $430 \pm 218$ & $94 \pm 10$ & $314 \pm 102$ & $60 \pm 29$ & $34 \pm 24$ \\
     $SA\mathcal{P}_\sigma^\text{SA}$ & $\mathbf{72 \pm 19}$ & $371 \pm 112$ & $\mathbf{86 \pm 13}$ & $\mathbf{620 \pm 195}$ & $\mathbf{100 \pm 0}$ & $353 \pm 69$ & $\mathbf{87 \pm 23}$ & $\mathbf{90 \pm 38}$ \\
     $SA\mathcal{P}_\sigma^\text{A}$ & $13 \pm 7$ & $51 \pm 33$ & $82 \pm 14$ & $285 \pm 148$ & $97 \pm 7$ & $\mathbf{404 \pm 83}$ & $76 \pm 29$ & $25 \pm 20$ \\
     $A(\cdot|s_0)$ & $25 \pm 11$ & $120 \pm 65$ & $71 \pm 23$ & $494 \pm 181$ & $99 \pm 3$ & $395 \pm 94$ & $78 \pm 25$ & $61 \pm 32$ \\
     \bottomrule
\end{tabular}
\caption{Mean and standard deviation of reward and survival rates (\%) over 100 initial states with 8 trajectories sampled for each. As in~\citep{Diffuser} we used Model Predictive Path Integral (MPPI)~\citep{MPPI} as a baseline, but it struggled due to the long-horizon nature of our tasks. The \textcolor{tablegray}{data} row refers to our training data generated by \emph{closed-loop} RL or MPC policies, which is why their performance exceeds all the \emph{open-loop} models. 
Among all the models tested $SA\mathcal{P}^\text{SA}_\sigma$ consistently perform among the best.}
\label{tab: reward survival Q3}
\end{table*}

\subsection{Diffusion modality}\label{subsec: Q4}

Question~\ref{Q: diffusion on S, SA, A} on choosing the best prediction modality is motivated by the large literature embracing each one of these approaches as discussed in Section~\ref{subsec: litt planning}. We have already introduced models predicting states only and others generating both states and actions. The last category generates action sequences conditioned on the current state. By generating only actions, the resulting trajectory is admissible. As pointed out in \citep{DecisionDiffuser} actions are more volatile than states rendering action sequences harder to predict. However, this claim was refuted in action planning models such as \citep{D-MPC}. The action space is typically smaller than the state space, which also plays in favor of action prediction.

Table~\ref{tab: Q1} shows that the models predicting both states and actions generate trajectories closer to admissible than the ones predicting only states. However, in term of admissibility all models predicting actions are superior to the states-only ones since their predicted sequence of actions always yield an admissible trajectory.

D-MPC~\citep{D-MPC} is an offline model-based approach learning a dynamics model and a policy using two different diffusion models. Because our work assumes access to a dynamics simulator, a fair comparison with D-MPC should only compare its diffusion policy. We trained such a model generating sequences of actions conditioned on the initial state $s_0$ on our datasets using our DiT architecture under the acronym $A(\cdot|s_0)$.

\begin{table}[htb!]
    \centering
    \begin{tabular}{cccc}
        \toprule
         System & Model & survival \% ($\uparrow$) &  reward ($\uparrow$) \\ \midrule
         \multirow{3}{*}{Hopper} & $S\mathcal{P}_\sigma^\text{ref}$ & $\mathbf{88 \pm 16}$ & $\mathbf{519 \pm 130}$ \\
         & $SA\mathcal{P}_\sigma^\text{SA}$ & $72 \pm 19$ & $371 \pm 112$ \\
         & $A(\cdot |s_0)$ & $25 \pm 11$ & $120 \pm 65$\\[2mm]
         \hdashline \\[-2mm]
         \multirow{3}{*}{Walker} & $S\mathcal{P}_\sigma^\text{ref}$ & $63 \pm 23$ & $365 \pm 134$ \\
          & $SA\mathcal{P}_\sigma^\text{SA}$ & $\mathbf{86 \pm 13}$ & $\mathbf{620 \pm 195}$ \\
          & $A(\cdot |s_0)$ & $71 \pm 23$ & $494 \pm 181$ \\[2mm]
         \hdashline \\[-2mm]
         \multirow{2}{*}{HalfCheetah} & $SA\mathcal{P}_\sigma^\text{SA}$ & $\mathbf{100 \pm 0}$ & $353 \pm 69$ \\
          & $A(\cdot |s_0)$ & $99 \pm 3$ & $\mathbf{395 \pm 94}$\\[2mm]
         \hdashline \\[-2mm]
         \multirow{3}{*}{Quadcopter} & $S\mathcal{P}_\sigma^\text{ref}$ & $\mathbf{89 \pm 19}$ & $89 \pm 33$ \\
          & $SA\mathcal{P}_\sigma^\text{SA}$ & $87 \pm 23$ & $\mathbf{90 \pm 38}$ \\ 
          & $A(\cdot |s_0)$ & $78 \pm 25$ & $61 \pm 32$ \\ \bottomrule
    \end{tabular}
    \caption{Reward and survival comparison between S, SA, and A models averaged over 100 initial states and 8 trajectories generated for each.
    We could not evaluate state models for the HalfCheetah due to the lack of performant inverse dynamics model. Despite their lack of admissibility guarantees state models perform well on the Hopper and Quadcopter. State-action model $SA\mathcal{P}_\sigma^\text{SA}$ is consistently near the best while action models $A(\cdot|s_0)$ achieve inconsistent results.}
    \label{tab: reward survival Q4}
\end{table}

As shown in Table~\ref{tab: reward survival Unitree} the Action model performance can vary from worse to best on extremely similar environments. Except for the Hopper where $A(\cdot|s_0)$ performs poorly, it shows decent results on the other environments of Table~\ref{tab: reward survival Q4}.  Due to its inconsistency predicting only action sequences does not perform as well as state-action models like $SA\mathcal{P}_\sigma^\text{SA}$, which answers \ref{Q: diffusion on S, SA, A}.

\begin{table}[htb!]
    \centering
    \definecolor{tablegray}{RGB}{90, 98, 115}
    \addtolength{\tabcolsep}{-1pt}
    \begin{tabular}{cccc}
        \toprule
         System & Model & survival ($\uparrow$) &  reward ($\uparrow$) \\ \midrule
         \multirow{4}{*}{Unitree GO1} & $SA(\cdot|c)$ & $87 \pm 25$ & $653 \pm 408$ \\
          & $SA\mathcal{P}^\text{SA}_\sigma(\cdot|c)$ & $87 \pm 28$ & $638 \pm 409$ \\
          & $SA\mathcal{P}^\text{A}_\sigma(\cdot|c)$ & $\mathbf{100 \pm 0}$ & $818 \pm 466$ \\
         & $A(\cdot|c, s_0)$ & $99 \pm 3$ & $\mathbf{1081 \pm 495}$ \\[2mm]
         \hdashline \\[-2mm]
         \multirow{4}{*}{Unitree GO2} & $SA(\cdot|c)$ & $\mathbf{100 \pm 0}$ & $1039 \pm 122$ \\
          & $SA\mathcal{P}^\text{SA}_\sigma(\cdot|c)$ & $\mathbf{100 \pm 0}$ & $\mathbf{1080 \pm 35}$ \\
          & $SA\mathcal{P}^\text{A}_\sigma(\cdot|c)$ & $\mathbf{100 \pm 0}$ & $1074 \pm 53$ \\
          & $A(\cdot|c, s_0)$ & $\mathbf{100 \pm 0}$ & $911 \pm 217$ \\
         \bottomrule
    \end{tabular}
    \caption{Reward and survival rates (\%) $\pm std$ over 100 initial states $s_0$ with 64 samples generated per $s_0$. All these models predict actions implemented open-loop on the robots and conditioned on the velocity command $c$. We could not evaluate state models for the Unitree due to the lack of performant inverse dynamics model. The action only model $A(\cdot|c, s_0)$ ranks respectively best and worse on the GO1 and GO2, whereas state-action model $SA\mathcal{P}^\text{A}_\sigma(\cdot|c)$ performs consistently well on both robots.}
    \label{tab: reward survival Unitree}
\end{table}

\section{Hardware Experiments}\label{sec: hardware}

In this section, we evaluate DDAT on real-world hardware. Specifically, we demonstrate that DDAT can generate high-dimensional trajectories deployable on hardware in open-loop. Moreover, DDAT causes fewer failures than a baseline diffusion model and outperforms it in task completion.
We train both diffusion policies in the Unitree GO2 environment and generate trajectories walking forward for $10$ seconds. To evaluate walking behavior, we aim for the quadruped to pass a goal line $3$ meters in front of the start position while staying within a $1$ meter corridor. If the quadruped leaves the sidelines or begins to fall, we consider it a failure. For this evaluation, we sampled $3$ trajectories from each model and performed $10$ trials. As such, we performed a total of $30$ trials for each model, and a summary of the results is presented in Table~\ref{tab: hardware}.

\begin{table}[htb!]
    \definecolor{tablegray}{RGB}{90, 98, 115}
    \centering
    \addtolength{\tabcolsep}{-1pt}
    \begin{tabular}{cccc}
        \toprule
         Model & Success ($\uparrow$) & Exits ($\downarrow$) & Falls ($\downarrow$) \\
         \midrule
         $SA\mathcal{P}_\sigma^\text{SA}(\cdot, |c)$ & $\mathbf{23/30}$ & $\mathbf{5/30}$ & $\mathbf{2/30}$ \\
         $SA(\cdot, |c)$ & $10/30$ & $16/30$ & $4/30$ \\
         \bottomrule
    \end{tabular}
    \caption{Evaluation of $60$ walking hardware trials of trajectories generated by diffusion models on the Unitree GO2. If the quadruped ever left the prescribed region (Exits) or lost balance to the point of falling, we considered the runs as failures.}
    \label{tab: hardware}
\end{table}

Overall, we found that DDAT had consistently higher success rates compared to the vanilla diffusion policy. Specifically, DDAT was successful in more than twice as many trials as the diffusion policy baseline. The baseline would often generate actions causing the robot to slip or begin to fall, leading to either a `Fall Failure' or a recovery performed by the onboard GO2 computer. However, even after such a recovery, the quadruped would often be pointed in the wrong direction, leading to the robot to leave the prescribed corridor, making it a `Exit Region Failure'. The baseline trajectory would also occasionally cause the quadruped motors to vibrate and overwork themselves until they locked up, and the robot would begin to slouch in place. We also considered such scenarios as `Fall Failure'. In comparison, our DDAT trajectories were often smoother, commanding the Unitree GO2 to follow a periodic motion, leading it along a fast and straight path to the goal line at $3$ meters without much deviation. Nonetheless, as these are open loop trajectories, the stochastic nature of the physical hardware caused some of the trajectories to leave the region. Overall, DDAT's superior performance showcases the importance of developing dynamically feasible and admissible trajectories for real world deployments.

We also successfully deployed DDAT on a Unitree GO1 quadruped with open-loop trajectories generated by another one of our diffusion model $SA\mathcal{P}^\text{SA}_\sigma$. While walking $2$m forward, the robot deviated only $0.32$m in a lateral direction as shown in the additional materials.

\section{Limitations}\label{sec: limitations}

To verify the dynamical admissibility of trajectories we assumed access to a perfect simulator. This limitation could be removed by instead learning a dynamics model from the training data if a perfect simulator is not available. Using a learned dynamics model instead of a black-box would also make it differentiable, hence allowing to completely back-propagate gradients through all of our projectors and hopefully leading to better training.

Diffusion transformers~\citep{DiT} and diffusion models in general require large neural network architectures, resulting in computationally intensive training.

Another limitation shared by most diffusion policies is their slow and computationally intensive inference. Compared to the traditional DDPM~\citep{DDPM} models requiring hundreds of inference steps, our implementation only requires a fraction of the inference steps (we used $N = 5$) but the projections can become costly for long-horizon and large action spaces. Speed improvements can certainly be achieved following an approach similar to that of~\citep{DiffuseLoco}.

\section{Conclusion and Future Work}\label{sec: conclusion}

In this work we introduced DDAT models trained to generate open-loop dynamically admissible trajectories, while only using a black-box dynamics model. We proposed several projection approaches to enforce the respect of dynamics based on the diffusion modalities. We developed a novel diffusion architecture incorporating curriculum projections during both training and inference. We successfully deployed our models on a variety of robotic environments both in simulations and hardware.
We have found that combined predictions of states and actions is the best way to generate dynamically admissible trajectories of high quality to leverage both the planning capabilities of state predictions and the guaranteed admissibility of action planners.

We envision several avenues of future work. 
We plan on deploying our approach in a purely offline setting without dynamics simulator which we will learn from trajectories, possibly using a diffusion model as~\citep{D-MPC}.
We also want to make our approach faster to be run in closed-loop on real hardware.

\section*{Acknowledgments}

This work is supported in part by the National Science Foundation, under grants ECCS-2438314 CAREER Award, CNS-2423130, CCF-2423131, and by the Google-BAIR Commons Project Year 6.
This work is supported in part by the Robotics and AI Institute. K. Sreenath has financial interest in the Robotics and AI Institute. He and the company may benefit from the commercialization of the results of this research.

The authors are also grateful to John Viljoen for sharing the code of his MPC and quadcopter environment.

\bibliographystyle{plainnat}
\bibliography{references}

\appendices

\section{Projection of actuated states}\label{apx: proj actu}

Following Algorithm~\ref{alg: proj} the projection operator $\mathcal{P}$ of \eqref{eq: projection} operates on the entire state $s \in \mathbb{R}^n$. However, we can accelerate this optimization by leveraging our knowledge of system dynamics and only projection roughly half of the states. The state vector $s \in \mathbb{R}^n$ describing robots is typically composed of two halves: a position vector $x \in \mathbb{R}^{n_1}$ representing the position and orientation of the robot, and a velocity vector $v \in \mathbb{R}^{n_2}$ with $n_1 + n_2 \leq n$. Without knowing the black-box dynamics model of~\eqref{eq: nonlinear dynamics}, but only knowing the meaning of each state, we can easily obtain a relation like $\dot x = v$ or something a bit more complex if quaternions are involved. Yet, even in this case the time derivative formulas are known.

For instance, if we assume the state is $s_t = (x_t, v_t)$ and the simulator of~\eqref{eq: nonlinear dynamics} uses a timestep $dt$ and Euler explicit integrator we can write $x_{t+1} = x_t + dt\, v_t$. Thus, the next position $x_{t+1}$ is entirely determined by the current state $(x_t, v_t)$ and does not need to be part of the projection. Hence, we can replace~\eqref{eq: projection} with
\begin{equation}\label{eq: actu projection}
    \mathcal{P}\big(\tilde{s}_{t+1}, \mathcal{C}(s_t)\big) := \big(x_t + dt\, v_t\,,\ \underset{r\, \in\, \mathcal{R}}{\arg\min} \|\tilde{v}_{t+1} - r\| \big),
\end{equation}
where $\mathcal{C}(s_t)$ is the reachable set approximation of \eqref{eq: reachable set approx}.

We can adapt this process to other numerical integrators, like for instance the semi-implicit Euler integrator available in MuJoCo~\citep{mujoco}, where $x_{t+1} = x_t + dt\, v_{t+1}$, which can be directly replaced in \eqref{eq: actu projection}.
If the state vector also includes unactuated velocities, then they are constant in the reachable set (even with semi-implicit Euler) and can be excluded from the optimization. The main insight being to remove as many state components as possible from the projection to speed it up by leveraging known system dynamics. 

Note that we can invert the integrator equations to obtain $v_{t+1}$ as a function of $x_t$ instead and do the projection on the positions instead of the velocities. However, our experiments showed that the generated trajectories diverge faster with a position projection, while velocity and full-state projection yield the same trajectory quality with a speed gain for the velocity-only projection.

\section{Inverse Dynamics}\label{apx: inverse dynamics}

\begin{algorithm}[htbp!]
\caption{Trajectory Inverse Dynamics}\label{alg: ID traj}
\begin{algorithmic}[1]
\Require predicted trajectory $\tilde{\tau} = \big\{s_0, \tilde{s}_1,... \tilde{s}_H\big\} \in \mathcal{S}^{H+1}$.
\vspace{2mm}
\State $\tau = \big\{ s_0 \big\}$ \Comment{initialize the admissible trajectory}
\For{$t=0$ to $H-1$}
\State $a_t = I\!D\big(s_t, \tilde{s}_{t+1}\big)$ \Comment{best action}
\State $s_{t+1} = f\big(s_t, a_t\big)$ \Comment{closest admissible next state}
\State $\tau \leftarrow \tau \cup \{s_{t+1}\}$ \Comment{append to $\tau$}
\EndFor
\State \textbf{return} $\tau$
\Ensure admissibility of $\tau$.
\end{algorithmic}
\end{algorithm}

Previous works in the literature such as \citep{DecisionDiffuser, romer2024safe} have claimed learning an inverse dynamics model using supervised learning. However, learned models cannot achieve a sufficient precision in their prediction leading to compounding errors and diverging trajectories over long-horizon predictions. Instead, we tried two different black-box optimization approaches to solve the inverse dynamics equation~\eqref{eq: ID}. Algorithm~\ref{alg: ID polytope} leverages the polytopic nature of the admissible action set $\mathcal{A}$ of most robotic systems to perform a convex optimization relying on control affine intuition detailed in Remark~\ref{rmk: control affine}.

\begin{remark}\label{rmk: control affine}
    Consider a control affine system $s_{t+1} = g(s_t) + h(s_t)a_t$. Let $\tilde{s}$ be the predicted next state of $s_t$. If there exist actions $a^i \in \mathcal{A}$ with corresponding successors $s^i = g(s_t) + h(s_t)a^i$ and convex coefficients $\lambda_i \in [0, 1]$ such that $\sum \lambda_i = 1$ and $\tilde{s} = \sum \lambda_i s^i$, then $\tilde{s}$ is reachable with action $\tilde{a} := \sum \lambda_i a^i$.
    Indeed,
    \begin{align*}
        \tilde{s} &= \sum \lambda_i s^i = \sum \lambda_i g(s_t) + \lambda_i h(s_t) a^i \\
        &= g(s_t)\sum \lambda_i + h(s_t) \sum \lambda_i a^i = g(s_t) + h(s_t) \tilde{a}.
    \end{align*}
\end{remark}

Recall that the vertices of the admissible action set $\mathcal{A}$ are denoted as $\mathcal{V}\big(\mathcal{A}\big) = \big\{v_1, ..., v_m\big\}$. Algorithm~\ref{alg: ID polytope} stops when reaching some precision threshold $\varepsilon <\!\!< 1$ or maximal number of iterations $N$. The size of the action space searched with convex optimization \eqref{eq: convex optim} decreases by a factor $\delta < 1$, typically $\delta = 0.5$.

\begin{algorithm}[htb!]
\caption{Inverse Dynamics by Polytopic Optimization} \label{alg: ID polytope}
\begin{algorithmic}[1]
\Require current state $s_t \in \mathcal{S}$, predicted next state $\tilde{s}_{t+1} \in \mathcal{S}$.
\vspace{-2mm}
\State $a = \frac{1}{m} \sum_{i=1}^m v_i$ \Comment{initial guess, if not provided}
\State $s_{t+1} = f(s_t, a)$ \Comment{admissible next state}
\State $n = 0$
\While{$n < N$ \textbf{and} $\|\tilde{s}_{t+1} - s_{t+1}\| > \varepsilon$}
\For{$i = 1$ to $m$}
\State $\hat{v}_i = a + \delta \big(v_i - \frac{1}{m}\sum v_i \big)$ \Comment{extremal actions}
\State $s^i = f(s_t, \hat{v}_i)$ \Comment{extremal next states}
\EndFor
\State $\lambda^* = $ solve~\eqref{eq: convex optim}
\State $a = \sum_{i = 1}^{m} \lambda_i^* \hat{v}_i$ \Comment{update the action}
\State $s_{t+1} = f(s_t, a)$ \Comment{admissible next state}
\State $n \leftarrow n+1$
\EndWhile
\Ensure Admissible next state $s_{t+1}$ is as close as possible to prediction $\tilde{s}_{t+1}$
\end{algorithmic}
\end{algorithm}

While Algorithm~\ref{alg: ID polytope} performs very well on the Hopper and Walker2D, it cannot achieve a sufficient precision on the HalfCheetah to prevent trajectories from diverging when applied on an admissible trajectory.

The second approach we implemented for the inverse dynamics is a typical black-box optimization detailed in Algorithm~\ref{alg: ID black box}. We randomly sample actions until finding one leading to a next state $s'$ closer to the prediction $\tilde{s}$. Then, we perform a linesearch along the direction of improvement to further reduce this distance to $\tilde{s}$.

\begin{algorithm}[htbp!]
\caption{Inverse Dynamics by Black-Box Optimization} \label{alg: ID black box}
\begin{algorithmic}[1]
\Require current state $s_t \in \mathcal{S}$, predicted next state $\tilde{s}_{t+1} \in \mathcal{S}$
\vspace{2mm}
\State $a \sim \mathcal{U}\big(\mathcal{A}\big)$ \Comment{initial guess, if not provided}
\State $s_{t+1} = f(s_t, a)$ \Comment{admissible next state}
\State $n = 0$
\While{$n < N$ \textbf{and} $\|\tilde{s}_{t+1} - s_{t+1}\| > \varepsilon$}
\State $\delta a \sim \mathcal{N}(0, \sigma)$ with $\sigma$ proportional to $\|\tilde{s} - s'\|$
\If{$\|\tilde{s} - s'\| > \|\tilde{s} - f(s, a + \delta a)\|$}
\State $a \leftarrow a + \delta a\, \underset{\alpha\, \geq\, 0}{\arg\min}\big\{ \big\|\tilde{s} - f(s, a + \delta a\, \alpha)\big\| \big\}$ \hspace{40mm} \Comment{linesearch along $\delta a$}
\State $s_{t+1} = f(s_t, a)$ \Comment{admissible next state}
\EndIf
\State $n \leftarrow n+1$
\EndWhile
\Ensure Admissible next state $s_{t+1}$ is as close as possible to prediction $\tilde{s}_{t+1}$
\end{algorithmic}
\end{algorithm}

We use Algorithm~\ref{alg: ID polytope} to provide an accurate first guess to Algorithm~\ref{alg: ID black box} which will get the desired precision. However, even this combination of algorithms is not sufficient to reach the desired precision for the quadcopter and the Unitree GO1 simulation. The latter is due to the large dimension of the action space, which requires $2^{12} = 4096$ vertices for Algorithm~\ref{alg: ID polytope}.

\section{Diffusion implementation details}\label{apx: diffusion}

For our diffusion model, we use most of the values from \citep{karras2022elucidating} as was done in the DiT implementation of \citep{dong2024aligndiff}. Table~\ref{tab: Env & DiT} summarizes the key parameters of each environment along with the DiT parameters used.

\begin{table*}[htb!]
    \definecolor{tablegray}{RGB}{90, 98, 115}
    \centering
    \begin{tabular}{cccccccccc}
        \toprule
         & \multicolumn{3}{c}{Environment parameters} & \multicolumn{6}{c}{DiT parameters} \\
         \cmidrule(lr){1-1} \cmidrule(lr){2-4} \cmidrule(lr){5-10}
         Robot & \begin{tabular}{@{}c@{}}number\\of states\end{tabular} & \begin{tabular}{@{}c@{}}number\\of actions\end{tabular} & \begin{tabular}{@{}c@{}}prediction\\horizon\end{tabular} & width & depth & \begin{tabular}{@{}c@{}}number\\of heads\end{tabular} & conditioned & \begin{tabular}{@{}c@{}}prediction\\modality\end{tabular} & \begin{tabular}{@{}c@{}}number\\of parameters\end{tabular} \\
         \cmidrule(lr){1-1} \cmidrule(lr){2-4} \cmidrule(lr){5-10}
         Hopper      & $12$ & $3$ & $300$ & $64$  & $3$ & $4$ & no       & S  &   $238,284$ \\
         Hopper      & $12$ & $3$ & $300$ & $64$  & $3$ & $4$ & no       & SA &   $238,671$ \\
         Hopper      & $12$ & $3$ & $300$ & $64$  & $3$ & $4$ & on $s_0$ & A  &   $421,507$ \\[1mm]
         Walker2d    & $18$ & $6$ & $300$ & $256$ & $4$ & $3$ & no       & S  & $3,757,842$ \\
         Walker2d    & $18$ & $6$ & $300$ & $256$ & $4$ & $4$ & no       & SA & $4,944,408$ \\
         Walker2d    & $18$ & $6$ & $300$ & $256$ & $4$ & $3$ & on $s_0$ & A  & $4,451,590$ \\[1mm]
         HalfCheetah & $18$ & $6$ & $200$ & $256$ & $4$ & $4$ & no       & SA & $4,944,408$ \\
         HalfCheetah & $18$ & $6$ & $200$ & $256$ & $4$ & $3$ & on $s_0$ & A  & $4,451,590$ \\[1mm]
         Quadcopter  & $17$ & $4$ & $200$ & $256$ & $4$ & $4$ & no       & S  & $4,940,817$ \\
         Quadcopter  & $17$ & $4$ & $200$ & $256$ & $4$ & $4$ & no       & SA & $4,942,869$ \\
         Quadcopter  & $17$ & $4$ & $200$ & $256$ & $4$ & $4$ & on $s_0$ & A  & $5,596,676$ \\[1mm]
         Unitree GO1 & $37$ & $12$& $500$ & $256$ & $4$ & $6$ & on $cmd$ & SA & $7,490,353$ \\
         Unitree GO1 & $37$ & $12$& $500$ & $256$ & $4$ & $6$ & on $cmd$, $s_0$ & A & $8,892,172$ \\ [1mm]
         Unitree GO2 & $37$ & $12$& $500$ & $256$ & $4$ & $6$ & on $cmd$ & SA & $7,490,353$ \\
         Unitree GO2 & $37$ & $12$& $500$ & $256$ & $4$ & $6$ & on $cmd$, $s_0$ & A & $8,892,172$ \\ \bottomrule
    \end{tabular}
    \caption{Environments and DiT parameters. Each DiT model generates either sequences of states "S", states and actions "SA", or only actions "A". The Unitree models are all conditioned on the velocity command $cmd$.}
    \label{tab: Env & DiT}
\end{table*}

During training we sample noise levels with $\ln(\sigma) \sim \mathcal{N}(p_m, p_s^2)$ with $p_m = -1.2$ and $p_s = 1.2$ as in~\citep{karras2022elucidating}.

For inference we select $N = 5$ denoising steps with noise scales $\sigma_0 = 80$, $\sigma_{N-1} = 0.002$ with 
\begin{equation*}
    \sigma_{i<N} = \big( \sigma_{0}^{1/\rho} + \frac{i}{N-1}(\sigma_{N-1}^{1/\rho} - \sigma_{0}^ {1/\rho}) \big)^\rho,
\end{equation*}
$\sigma_N = 0$ and $\rho = 7$.
Our projection curriculum~\eqref{eq: proj scheduling} uses $\sigma_\text{max} = 0.2$ and $\sigma_\text{min} = 0.0021$, so that $\sigma_\text{min} > \sigma_{N-1}$ to have projections occur before the end of inference and complete projections.

We use a first-order deterministic sampling corresponding to Algorithm~1 of \citep{karras2022elucidating} without its lines 6 to 8. Using the recommended values in \citep{karras2022elucidating} of $t_i = \sigma_i = \sigma(t_i)$, $\dot\sigma(t_i) = 1$, $s(t_i) = 1$ and $\dot s(t_i) = 0$ we can simplify lines 4 and 5 of Algorithm~1 of \citep{karras2022elucidating} as follows
\begin{align*}
    \mathbf{d}_i &= \left(\frac{\dot\sigma(t_i)}{\sigma(t_i)} + \frac{\dot s(t_i)}{s(t_i)} \right) \mathbf{x}_i - \frac{\dot\sigma(t_i) s(t_i)}{\sigma(t_i)} D_\theta \left( \frac{\mathbf{x}_i}{s(t_i)}; \sigma(t_i) \right) \\
    &= \left(\frac{1}{\sigma_i} + \frac{0}{1} \right) \mathbf{x}_i - \frac{1}{\sigma_i} D_\theta \left( \mathbf{x}_i; \sigma_i \right) = \frac{\mathbf{x}_i - D_\theta \left( \mathbf{x}_i; \sigma_i \right)}{\sigma_i}.
\end{align*}
Then, line 5 becomes:
\begin{align}
    \mathbf{x}_{i+1} &= \mathbf{x}_i + (t_{i+1} - t_i) \mathbf{d}_i \nonumber \\
    &= \mathbf{x}_i + (\sigma_{i+1} - \sigma_i) \frac{\mathbf{x}_i - D_\theta \left( \mathbf{x}_i; \sigma_i \right)}{\sigma_i} \nonumber \\
    &= \frac{\sigma_{i+1}}{\sigma_i}\mathbf{x}_i + \left(1 - \frac{\sigma_{i+1}}{\sigma_i}\right)D_\theta \left( \mathbf{x}_i; \sigma_i \right) \nonumber \\
    &:= S_\theta(\mathbf{x}_i; \sigma_i, \sigma_{i+1}).
\end{align}
Then, the inference algorithm applied to trajectories $\tau$ reduces to Algorithm~\ref{alg: basic sampling}.

\begin{algorithm}[htbp!]
\caption{$1^{st}$ order deterministic sampling from~\citep{karras2022elucidating}}
\label{alg: basic sampling}
\begin{algorithmic}[1]
    \State $\tau_0 \sim \mathcal{N}(0, \sigma_0^2 I)$ \Comment{Sample random sequence}
    \For{$i \in \{0, ..., N-1\}$}
        \State $\tau_{i+1} \leftarrow S_\theta(\tau_i; \sigma_i, \sigma_{i+1})$
    \EndFor
    \textbf{return} $\tau_0$ \Comment{Noise-free sample}
\end{algorithmic}
\end{algorithm}

\section{Experiments details}\label{apx: experiments}

To be able to certify the admissibility of generated trajectories, we first need to be able to reproduce admissible trajectories. More specifically, starting from the same initial state, applying the same sequence of action in open-loop should generate the given sequence of states. However, while MuJoCo simulation~\citep{mujoco} is deterministic, trajectories are not exactly reproducible due to compounding errors arising from numerical computation differences across versions. Thus, we could not use the standard D4RL datasets~\citep{d4rl} and instead created our own datasets of reproducible trajectories. We trained PPO policies~\citep{PPO} to generate trajectories for the Hopper, Walker, and HalfCheetah~\citep{Gym}. We used model predictive control (MPC) to produce challenging trajectories for a quadcopter~\citep{viljoen2024differentiable} detailed in Appendix~\ref{apx: quad}. Finally, for the complex locomotion policies for Unitree GO1 and GO2, we used the massively parallelized Brax framework~\citep{brax} and MuJoCo playground~\citep{mujoco_playground} to train PPO policies~\cite{PPO}.

\subsection{HalfCheetah}\label{apx: cheetah}

Due to numerical instabilities the inverse dynamics of the HalfCheetah must be very precise to prevent trajectory divergence. Table~\ref{tab: Q1} shows that the inverse dynamics need to achieve $10^{-12}$ precision at each step to cap the cumulative divergence at $10^{-2}$ whereas the Hopper and Walker obtain similar CAE with much worse single step precision SAE of order $10^{-5}$. To prevent further worsening of the cumulative error of the HalfCheetah's inverse dynamics we restricted the prediction horizon to 200 steps, while the Walker and Hopper have 300 steps.

\subsection{Quadcopter}\label{apx: quad}

While the other environments are fairly standard we used a custom quadcopter environment based on~\citep{viljoen2024differentiable}. All other robots rely on ground contact to move forward, making their dynamics discontinuous which can lead to nonconvex reachable sets~\citep{azhmyakov2007convexity} and infeasible projected trajectories as discussed in Remark~\ref{rmk: inadmissible proj}. To illustrate our approach on a robot with smooth dynamics we chose the quadcopter.

\begin{figure*}[htb!]
    \centering
    \begin{subfigure}[]{0.32\textwidth}
        \includegraphics[scale=0.41]{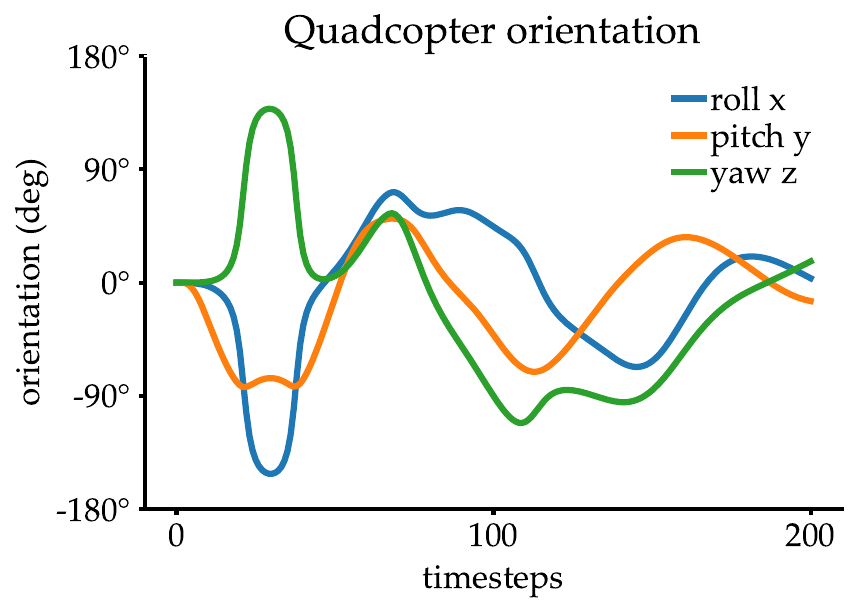}
        \caption{Orientation of the quadcopter}
        \label{fig: quad orientation}
    \end{subfigure}\hfill
    \begin{subfigure}[]{0.32\textwidth}
        \includegraphics[scale=0.42]{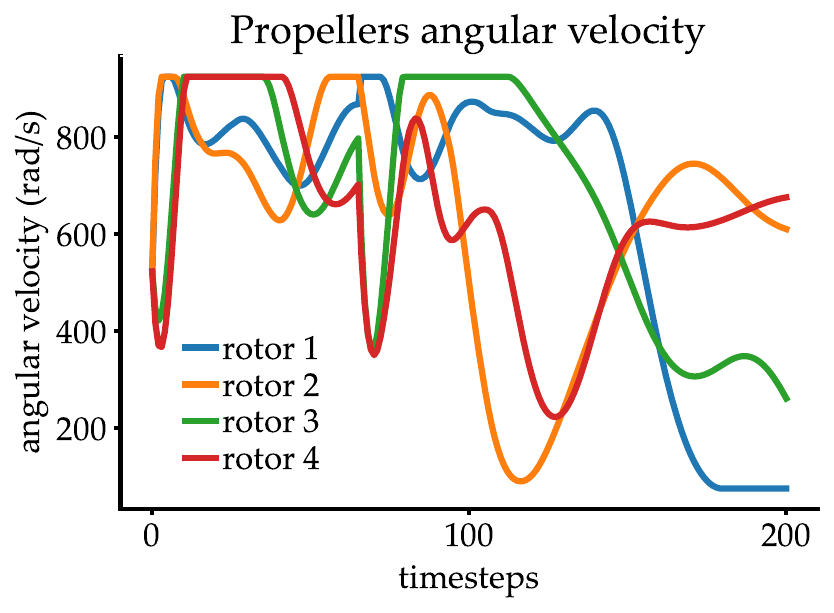}
        \caption{Angular velocity of the propellers}
        \label{fig: quad prop}
    \end{subfigure}\hfill
    \begin{subfigure}[]{0.345\textwidth}
        \vspace{5mm}
        \includegraphics[scale=0.45]{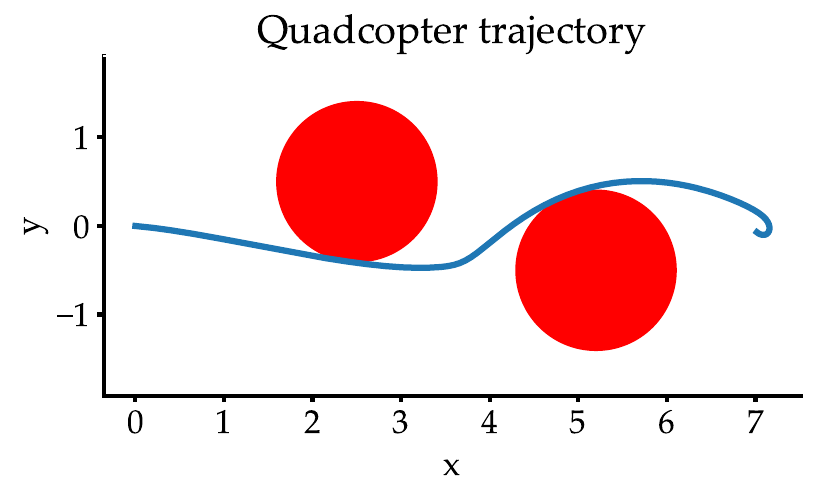}
        \caption{Quadcopter \textcolor{blue!70}{trajectory} between \textcolor{red!80}{obstacles}.}
        \label{fig: quad slalom}
    \end{subfigure}
    \vspace{5mm}
    \caption{Quadcopter reference trajectory generated using model predictive control (MPC)~\citep{viljoen2024differentiable}.}
    \label{fig: Quad}
\end{figure*}

Its state is composed of the position of its center of mass $x, y, z$, quaternion orientation $q_0, q_1, q_2, q_3$, linear velocity of the center of mass $v_x, v_y, v_z$, angular velocity $p, q, r$ and angular velocities of the propellers $w_1, w_2, w_3, w_4$. The actions are the angular acceleration of the propellers $\dot w_1, \dot w_2, \dot w_3, \dot w_4$.
The quadcopter starts around the origin at hover and is tasked with reaching position $x = 7$ in minimum time thanks to a model predictive controller (MPC) implemented in CasADi~\citep{Casadi}. The quadcopter must dodge two circular obstacles shown in Fig.~\ref{fig: quad slalom}. To reach the target in minimal time while dodging the obstacles, the MPC takes full advantage of all the degrees of freedom of the quadcopter as seen in Fig.~\ref{fig: quad orientation} and pushes the propellers to their maximal speed as shown in Fig.~\ref{fig: quad prop}. 

We generated $1000$ trajectories starting from randomized initial states around the origin to create a training dataset. The MPC module used obstacles of radii $0.9$ and generated trajectories tangent to their boundaries. We reduced the difficulty for the diffusion models with obstacles of radii $0.7$ allowing more room in between them.

Inverse Dynamics Algorithms~\ref{alg: ID polytope} and \ref{alg: ID black box} could not achieve sufficient precision in a reasonable time.
The inverse dynamics solver of the quadcopter is using the inertia of the propellers to match their angular velocity. This model knowledge is only used to judge the performance of our trained models which do not have access to this data due to our black-box assumption.

\subsection{Unitree GO1}

For the training dataset, we trained a joystick policy that can follow random commands consisting of the linear velocity of the base $v_x$, $v_y$, and yaw velocity of base $r$ using MuJoCo playground~\citep{mujoco_playground}. We generated 1000 trajectories with randomized commands and train a diffusion model conditioned on commands. Each trajectory includes 500 steps. The state space consists of the position of the base $x, y, z$, quaternion orientation of the base $q_0, q_1, q_2, q_3$, linear velocity of the base $v_x, v_y, v_z$, linear acceleration of the base $a_x, a_y, a_z$, angle of each joint $\theta_1, \dots, \theta_{12}$, and angular velocity of each joint $\dot{\theta}_1, \dots, \dot{\theta}_{12}$. The action space is the target joint angles $\theta_1^*, \dots, \theta_{12}^*$ for the low-level PD controller. We did not use any inverse dynamics as Algorithms~\ref{alg: ID polytope} and \ref{alg: ID black box} could not achieve sufficient precision in a reasonable time, and the GO1 dynamics are a black-box since simulated with MuJoCo~\cite{mujoco_playground}.

\subsection{Unitree GO2}
As the Unitree GO2 simulation is not supported by MuJoCo playground~\citep{mujoco_playground}, we implemented a custom environment based on GO1 environment and MuJoCo menagerie~\citep{mujoco_menagerie}. We used the same state space and action space as the GO1 environment for the diffusion planner and trained it with 1000 trajectories with randomized commands.
Similar to the GO1 environment, we did not have access to any inverse dynamics as Algorithms~\ref{alg: ID polytope} and \ref{alg: ID black box} could not achieve sufficient precision in a reasonable time, and the GO2 dynamics are a black-box simulated with MuJoCo~\cite{mujoco}.

\subsection{Hardware settings}

We introduced detailed hardware settings of Unitree GO1 and GO2 environments here. For both models, we generate trajectories of walking straight with the command of $[v_x, v_y, r] = [1.0, 0.0, 0.0]$. Action space is the target joint angle for the low-level PD controller. DDAT generates $10$ seconds trajectories with a control frequency of $50$Hz, meaning that each trajectory consists of $500$ steps.

In the GO2 experiments, we deploy a $50$Hz policy at $500$Hz via a zero-order-hold of the reference signal for the $10$Hz in between new points. Our open-loop controller reads a diffusion-generated trajectory from a CSV file and sends a command message to the robot using the Unitree SDK API with a ROS2 framework~\cite{ros2}. Then, the target joint angle is translated to the appropriate PWM commands for each motor by the onboard ICs and the prescribed PD parameters $K_p = 35.0, K_d=0.5239$. Before each deployment, the quadruped was first commanded for $4$ seconds to reach the initial position used to generate the trajectories from the diffusion models, as an attempt to reduce the sim-to-real gap.

In the evaluation of GO2 experiments, we mark the rollout as a success if the robot reaches the goal line which is $3$ meters front from the start position without leaving the sidelines, which has $1$ meter distance from the start position. If the quadruped left this region or began to fall, we would consider this a failure. It is important to note that Unitree included a self-righting mechanism, so we consider large imbalances as falls in order to terminate before any hardware is damaged.

In the GO1 experiments, we deployed a policy using a legged control framework~\citep{leggedcontrol}, which enables sim-to-sim and sim-to-real validation of the robot controller. We also run our open-loop controller by sending a DDAT-generated sequence of actions to the robot.

\subsection{Computational overhead}

The iterative dynamics sampling and the CVXPY optimization~\cite{cvxpylayers2019} of projectors~\eqref{eq: projection} and \eqref{eq: ref proj} generate additional computational costs. Indeed, these approaches scales exponentially in the dimension of the action space. The action set $\mathcal{A}$ is typically a hypercube $[-1, 1]^m$ with $2^m$ vertices.
As shown on Table~\ref{tab: inference time}, the models with reachable set projections $S\mathcal{P}_\sigma$ and $S\mathcal{P}_\sigma^\text{ref}$ are much slower than the model without projection $S$. That is why we introduced models $SA\mathcal{P}_\sigma^\text{A}$ and $SA\mathcal{P}_\sigma^\text{SA}$ capable of scaling to much larger action sets (like the GO2).

\begin{table}[h!]
    \centering
    \begin{tabular}{ccccccc}
        \toprule
        $\backslash$ & $S$ \hspace{-1mm}& $S\mathcal{P}_\sigma$  \hspace{-2mm} & $S\mathcal{P}_\sigma^\text{ref}$  \hspace{-3mm} & $SA$ \hspace{-1mm} & $SA\mathcal{P}_\sigma^\text{A}$  \hspace{-2mm} &  $SA\mathcal{P}_\sigma^\text{SA}$ \hspace{-1mm}  \\
        \cmidrule(lr){1-1} \cmidrule(lr){2-4} \cmidrule(lr){5-7}
        Hopper & 0.1 & 28 & 7 &  0.2 & 0.5 & 1 \\
        Walker & 0.3 & 32 & 20 & 0.5 & 0.8 & 1.3 \\
        Quadcopter & 0.3 & 24 & 39 & 0.3 & 0.5 & 0.8 \\ 
        GO2 & - & - & - & 2.6 & 5.5 & 8.7 \\ \bottomrule
    \end{tabular}
    \caption{Inference time (in seconds) to generate 8 trajectories of 300 timesteps each on a cpu Intel Core i5-14400F 2.5GHz with 16GB of RAM. Our models $S\mathcal{P}_\sigma$ and $S\mathcal{P}_\sigma^\text{ref}$ are significantly slower than baseline $S$ due to their use of CVXPY projections and cannot scale up to the GO2. Our models $SA\mathcal{P}_\sigma^\text{A}$ and $SA\mathcal{P}_\sigma^\text{SA}$ are slower than the baseline $SA$ but scale better than the state models with CVXPY projections.}
    \label{tab: inference time}
\end{table}

We believe our models with projections can be optimized to run faster following the method of~\cite{DiffuseLoco} as for now no efforts were made to accelerate the computations. 
All the diffusion models were trained on a single Nvidia A100 gpu. Training took between a few hours for the Hopper to a full day for the Unitree GO2. We are now working on accelerating inference for closed-loop real time deployement.

\end{document}